# Abstract representational geometry supports inference in large language models

Yunan Zeng[1], Yuwang Wang[2]*

[1] College of Future Information Technology, Fudan University, Shanghai, 200433, China

[2] Beijing National Research Center for Information Science and Technology, Tsinghua University, Beijing, 100084, China

*Correspondence: wang-yuwang@tsinghua.edu.cn

**Abstract**

A defining feature of human intelligence is the ability to adapt to changing environments by inferring latent task structure from sparse observations. Neuroscientific research indicates that this capability relies on the hippocampus constructing abstract representations, expressed as low-dimensional, approximately orthogonal manifolds in neural state space[1]. However, the internal mechanisms of large language models (LLMs) remain largely opaque, making it unclear whether they form comparable abstract representations or instead rely on task-specific statistical regularities when performing comparable reasoning tasks. Here we adapt a contextual reversal-learning paradigm to a text-based setting and compare humans and LLMs at both the Behavioural and representational levels. We report that although LLMs exhibit generalizable reasoning less frequently than humans, when such inference occurs, their internal states exhibit abstract geometric structures that resemble those reported in the hippocampus. Notably, this representational geometry is not uniformly distributed but is organized hierarchically across model depth: whereas lower layers show early, stable encoding of stimulus identity, higher layers form a hippocampal-like functional band enriched for abstract context geometry associated with inference. Furthermore, complementary intervention experiments mechanistically implicate geometry in reasoning: task-sequence language modelling induces geometric disentanglement, whereas geometric regularization of higher layers increases the emergence of generalizable inference. Together, these findings establish abstract representational geometry as a mechanistic principle supporting inference in large language models.

## Introduction

Humans possess a remarkable capacity to adapt to dynamic environments. Central to this adaptability is structural generalization—the ability to infer latent variables (such as hidden contexts or rules) from sparse observations and transfer this abstract knowledge to novel situations. Unlike simple associative learning, which relies on surface-level correlations, this mechanism enables agents to extract the underlying structure of the world, thereby flexibly adapting to complex environmental changes[2–4]. As illustrated in Fig. 1a, humans can flexibly switch from the mouse scroll direction of the Windows operating system to the inverted direction in MacOS, or transition from left-hand traffic rules to right-hand traffic rules.

Seminal findings in systems neuroscience indicate that this capability relies on the hippocampus constructing low-dimensional "cognitive maps" of abstract conceptual spaces[5–8]. These maps constitute abstract representations, defined as high-level codes that factorize the latent structure of a task from immediate sensory details. Recent high-dimensional neural recordings have further elucidated the geometric nature of these representations: during reasoning, neural populations spontaneously organize into disentangled, orthogonal manifolds[1,9–11]. In this geometric schema, known as factorized representation[12,13], abstract context variables and sensory variables are encoded along mutually orthogonal axes. This enables the brain to recombine known structures with novel sensory inputs, achieving rapid adaptation to changing environments[14,15]. Crucially, from a normative perspective, the formation of such factorized neural manifolds is increasingly understood as an efficient representational strategy necessitated by the brain's stringent biophysical limitations on energy expenditure and axonal wiring[16–18].

The structural necessity of biological intelligence raises a profound question regarding large language models (LLMs). Although they are not subject to the same biophysical constraints, LLMs rely instead on scaling laws—achieving increasingly complex reasoning behaviours through the systematic expansion of network parameters and massive data training[19–23]. This contrast raises the question of whether such geometric representations are substrate-general: Are the capabilities of LLMs driven entirely by superficial statistical fitting[24,25]? Or do LLMs spontaneously replicate this classical geometric decoupling as an emergent property of massive scale, suggesting that intelligence, regardless of its substrate, may converge on related mathematical structures?

Prevailing efforts in mechanistic interpretability have largely focused on localizing conceptual units, employing techniques such as Sparse Autoencoders (SAEs) to decode the "atoms" of static semantic features[26,27] or mapping concept directions under the linear representation hypothesis[28–30]. Yet, while these methods successfully disentangle semantic primitives, they are not designed to fully capture the distributed, population-level dynamics intrinsic to logical reasoning[31,32]. Biological studies suggest that flexible rule-guided computation is not implemented by isolated units alone, but emerges from coordinated population-level organization in neural state space[1,33,34]. Characterizing this population-level geometry—the dynamic "shape" of thought—offers a novel lens for peering into the black box of reasoning in LLMs, providing a mechanistic framework to investigate whether such geometric statistical properties emerge in distributed representations and how they might support abstract structural generalization.

Here, we adapted the contextual reversal learning paradigm[35] to representative LLMs to determine whether they possess the capacity for rapid adaptation and if it is underpinned by geometry analogous to population-level codes observed in neuroscience. To minimize semantic priors, we designed a symbolic task variant using nonsense stimuli to reduce semantic cues, thereby compelling the model to infer latent variables solely from the structural patterns of interaction history. To ensure the robustness and universality of our findings, we conducted a comprehensive analysis across two widely used open-weight model families: LLaMA[36] and Qwen[37], alongside representative proprietary models (e.g., GPT-5.1[22], Claude-4.5-Sonnet[38], etc.). This selection spans an order of magnitude in parameter scale, enabling us to decouple architectural specificity from fundamental scaling properties. For the open-weight models, we probed their internal representations and adopted two key metrics derived from systems neuroscience: Cross-Condition Generalization Performance (CCGP)[9], which measures the cross-condition generalizability of abstract variables, and Parallelism Score (PS)[9] , which evaluates the consistency of coding directions across task conditions.

Leveraging this framework, our study reveals three linked findings, moving from representational geometry to architectural localization and interventional evidence:

First, at the phenomenological level, we identified a striking representational convergence between biological and artificial intelligence. Despite the fact that humans engage in reasoning significantly more frequently than current models, we found that when LLMs successfully infer latent variables, their internal states organize into orthogonal, disentangled manifolds resembling those reported in the human hippocampus. This comparison moves beyond neural-response similarity during language understanding[39,40] to a deeper neural-computational level: internal states of LLMs exhibit a representational geometry analogous to neural population codes for inference, factorizing latent task structure from sensory content. This provides a concrete geometric criterion for distinguishing structural inference from rote stimulus–response association in language models.

Second, at the architectural level, we discovered a spontaneous functional specificity within the Transformer that is reminiscent of hierarchical organization in biological systems. We found that inference-related geometry is not uniformly distributed across depth. Lower layers already encode stimulus identity with high fidelity, whereas middle-to-higher layers are selectively enriched for abstract context geometry, providing a hippocampal-like computational locus in which latent task context can be maintained, factorized from stimulus identity, and used to support generalizable rule updating.

Third, we provide two complementary lines of interventional evidence indicating that representational geometry is not merely correlated with inference but functionally involved in it. Fine-tuning on task sequences showed that learning to model task structure naturally induces geometric disentanglement—evidenced by a ~33% enhancement in parallelism (PS) and a ~13% gain in generalization (CCGP). Conversely, direct geometric regularization of higher layers increased the prevalence of generalizable reasoning by 2.7-fold, strengthening the evidence that abstract geometry plays a functional role in structural generalization.

### Text-based Reversal Learning Task

We designed a text-based reversal-learning task (Fig. 1b) to evaluate whether humans and LLMs can infer latent task structure from sparse feedback under implicit context switches. In each trial, the subject/model was presented with one of four pseudoword stimuli (for example, "baf", "teq", "zir" or "dru"), selected a binary response (left or right), and received immediate outcome feedback. Correct responses yielded one of two non-zero outcomes ($25 or $5), whereas incorrect responses yielded $0. The task contained two latent contexts (Context 1, Context 2), each defined by a fixed stimulus-response-outcome (S-R-O) mapping. Across contexts, response mappings were globally reversed: the response that was correct for a given stimulus in one context became incorrect in the other, while the alternative response became correct. Because the active context switched without an explicit cue, successful performance required the agent to infer the current latent context from the sequence of stimuli, responses and outcomes. This design distinguishes structural inference from stimulus-by-stimulus relearning. After a context switch, an agent that simply memorizes individual stimulus–response associations will initially repeat the previously correct response and must relearn each stimulus separately. In contrast, an agent that infers a latent context switch from a single negative feedback event can reverse the entire rule set and respond correctly to other stimuli before observing them in the new context. Thus, correct responses to previously unseen stimuli in the new context provide a Behavioural signature of generalizable inference rather than local associative updating.

To establish a human baseline, we implemented the task as a web-based Behavioural experiment using Amazon Mechanical Turk. Participants viewed pseudoword stimuli, made left/right choices, and received immediate reward feedback, allowing us to measure trial-by-trial adaptation to unannounced S-R-O reversals. To evaluate LLMs under the same task structure, we converted each session into an autoregressive text sequence (Fig. 1c). The model received the cumulative history of completed trials together with the current stimulus and predicted the next response over $\{\text{left, right}\}$. Specifically, the task history is serialized as a continuous text stream of alternating stimulus ($s$), action ($a$), and outcome ($o$) tokens. At each time step $t$, the model receives a context sequence $x_t = [s_1, a_1, o_1, \cdots, s_{t-1}, a_{t-1}, o_{t-1}, s_t]$, which integrates all past interactions with the current stimulus $s_t$. The model utilizes this context to predict the corresponding action $\hat{a}_t$. Subsequently, the system appends the actual action and environmental outcome to the sequence, establishing a closed-loop feedback mechanism that drives continuous in-context inference. This text-based implementation preserved the same latent context structure and feedback logic used in the human task while allowing hidden-state representations to be extracted for subsequent geometric analyses. Specific experimental procedures, data exclusion protocols, and grouping criteria are detailed in the Methods section.

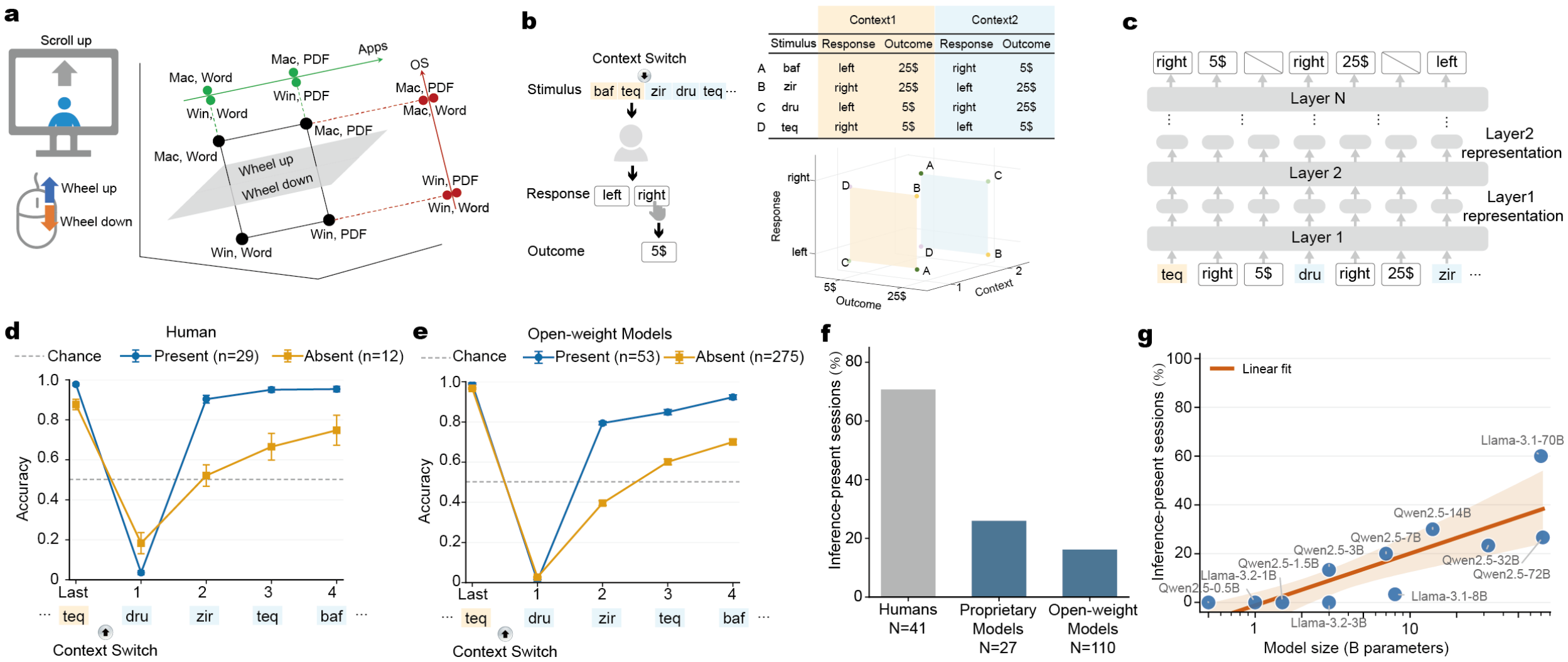

**Fig. 1 | The structural generalization task and Behavioural convergence between humans and large language models. a**, Conceptual illustration of structural generalization. Humans can disentangle abstract rules (e.g., scroll direction in different OS) from specific sensory contents, enabling rapid transfer across contexts. The right panel illustrates the idealized factorized geometry (parallelogram) of such representations. **b**, The Contextual Reversal Learning task design. Left: Task flow. The agent receives a pseudoword stimulus (e.g., "baf"), makes a binary response ("left"/"right"), and receives an outcome ($5/$25). Top right: Stimulus-response maps. A latent context (Context 1/Context 2) governs the mapping between stimulus-response pairs and outcomes; rules reverse upon context switching. Bottom right: Theoretical disentangled geometry where stimulus, response, and context are encoded along orthogonal axes. **c**, Network architecture. The task is modeled as an autoregressive sequence prediction problem using an interleaved input sequence of stimuli, responses, and outcomes. Hidden representations are taken from transformer layers at specific token positions within each trial (e.g., before response tokens), yielding embeddings for downstream representational-geometry analyses. **d-e**, Behaviour aligned to context switches in humans (d) and open-weight language models (e). Accuracy (mean ± s.e.m. across sessions; n shown in legend) is plotted for trials surrounding each switch, capturing the initial post-switch error and the subsequent inference trials on the remaining stimuli. Sessions are classified as inference-present or inference-absent based on above-chance performance on the first inference trial following a switch (binomial test); dashed line indicates chance level (0.5). **f**, Inference-present session rate across humans and language models (proprietary and open-weight). Bars show the percentage of sessions classified as inference-present (present / (present + absent); excluded sessions omitted). Sample sizes (total sessions) are indicated under each group (Humans, n = 41; Proprietary, n = 27; Open-weight, n = 328). **g**, Inference-present session rate increases with model size. Each point denotes one base model; the y-axis reports the percentage of inference-present sessions (as defined in the preceding panel). Model size is given in billions of parameters (log scale). Orange line, linear fit versus log10(parameter count); shaded band, 95% bootstrap confidence interval. (Pearson $r = 0.827$, $p = 0.002$; Spearman $\rho = 0.888$, $p < 1e{-3}$; computed on log10 parameter count).

## Behaviours of Humans and LLMs

To discern whether LLMs possess abstract reasoning capabilities or rely on simple associative learning (i.e., the gradual accumulation of specific stimulus-response mappings), we analyzed their Behavioural dynamics in the contextual reversal learning task. A recent landmark study by Courellis et al.[1] established a Behavioural hallmark for this distinction while investigating the emergence of abstract representations in the human hippocampus. Adopting their quantitative criteria, we reasoned that genuine inference would prompt a global update of the latent context following a single error, enabling accurate responses to stimuli unseen in the new context. Accordingly, we categorized sessions as "Inference Present" if accuracy on the trial immediately following the initial context-switch error significantly exceeded chance (binomial test, $P < 0.05$), and "Inference Absent" otherwise.

Under the "Inference Present" condition, humans and LLMs exhibited a strikingly isomorphic performance trajectory (Fig. 1d and e, blue curve), underscoring a convergent computational mechanism for rapid adaptation. Specifically, after an unprompted context switch (Trial 1), accuracy for both groups dropped to near 0%. This immediate failure confirms that both biological and artificial agents initially persisted in applying the obsolete "stimulus-response" mapping, demonstrating a shared reliance on the established latent context prior to receiving an explicit error signal. Crucially, however, upon receiving a single instance of negative feedback in this initial trial, accuracy in the subsequent trial (Trial 2) showed a highly aligned step-function increase, with humans recovering to >90% and LLMs to approximately 80%. This rapid, single-shot recovery demonstrates that a single error signal is sufficient to trigger a global Behavioural reversal for both humans and LLMs. Consequently, such a shared dynamic provides compelling evidence that neither system relearns associations stimulus-by-stimulus, but instead executes a unified, structural update to their internal task model. In contrast, the "Inference Absent" condition was characterized by a slow, incremental recovery curve (Fig. 1e, yellow line), consistent with standard reinforcement learning dynamics where policy updates occur slowly through repeated exposure to stimuli.

In summary, both humans and LLMs exhibited two behavioural regimes within the same task: the coexistence of rapid structural inference in "Inference Present" sessions and the gradual updating of stimulus-response associations in "Inference Absent" sessions. This significant qualitative convergence in Behavioural dynamics suggests that both possess mechanisms to switch flexibly between hypothesis-driven inference and simple associative strategies.

**Human Advantage in Inference Prevalence and Model Scaling Laws**

Although LLMs and humans exhibited similar Behavioural dynamics within both the "Inference Present" and "Inference Absent" modes, we observed a pronounced quantitative gap in their overall distribution: the proportion of LLM sessions classified as "Inference Present" was significantly lower than that of humans (Fig. 1f).

Within the population of LLMs, a clear performance hierarchy emerged: Proprietary Models (including Claude-Sonnet-4.5, Gemini-2.5-Flash, GPT-5.1, and Grok-4-Fast-Non-Reasoning) generally exhibited a higher frequency of reasoning than open-weight models (such as the Llama-3.1 and Qwen-2.5 series). This difference may reflect several factors, including model scale, training data, instruction tuning and alignment procedures such as RLHF[41], which may reshapes their inductive biases from surface-level heuristics to abstract structural priors.

Furthermore, within the open-weights model family, this reasoning capability approximately followed a power-law scaling relationship with model parameter count (Fig. 1g), which is consistent with foundational neural scaling laws[42]. We speculate that increasing model scale expands the representational capacity available for organizing task variables, making it easier to separate latent context from stimulus-specific information and to form a reusable context axis for rapid rule updating after sparse feedback.

**Representation Extraction and Analysis of LLMs**

To relate behavioural inference to model-internal representations, we first defined a sampling scheme over model depth, token position, and trial stage for extracting hidden states during the task. Specifically, for each trial, we wrote the task history into the model context and extracted hidden-state vectors at the token position immediately preceding response generation (Fig. 1c). Across model depth, we excluded the first layer (embedding output) and the last layer, using only the intermediate layer representations of the Transformer to reduce the direct influence of static token encoding and output readout constraints on geometric structure. Temporally, we sampled hidden states at two stages of each trial: a baseline stage immediately before the current stimulus was appended to the context, and a stimulus stage after the current stimulus had been appended, when the model was preparing to generate a response. Analogous to baseline and stimulus periods in neural recordings, the baseline stage indexes the carried-over task state, whereas the stimulus stage captures its integration with the current input for response selection.

Having defined the sampling scheme, we next organized trial-level hidden states into a condition-level geometry for downstream decoding and abstraction analyses. At each time point (baseline and stimulus), we aggregated the hidden representations of the trials according to the 2×2×2 combinations of the variables context, response, and outcome, thereby obtaining a representational geometry defined by the 8 points corresponding to 8 discrete conditions (see Methods for sampling and preprocessing details). Within these 8 conditions, we enumerated all 35 balanced dichotomies (4 vs 4) and focused on four dichotomies most directly related to task inference: context, response, outcome, and stim pair.

Finally, to rigorously determine whether this constructed geometry harbors abstract representations, we quantified the spatial arrangement of task variables using three complementary metrics derived from systems neuroscience. Intuitively, an ideal abstract encoding organizes key task variables in a low-dimensional, separable, and cross-background consistent geometry (Fig. 1b). Accordingly, we evaluate these features using Cross-Condition Generalization Performance (CCGP) to directly assess the cross-condition generalizability and separability of a variable, Parallelism Score (PS) to geometrically verify whether coding directions remain consistent and parallel across disparate backgrounds, and decoding accuracy to establish a baseline for linear separability, thereby precluding the possibility that such abstract structure is merely an artifact of high-dimensional space. By evaluating these three metrics in parallel under the same sampling and statistical framework, we robustly answer the core question of this paper: Do LLMs form abstract geometric structures supporting inference during implicit context switches, and how do these structures manifest at different time points in the task process?

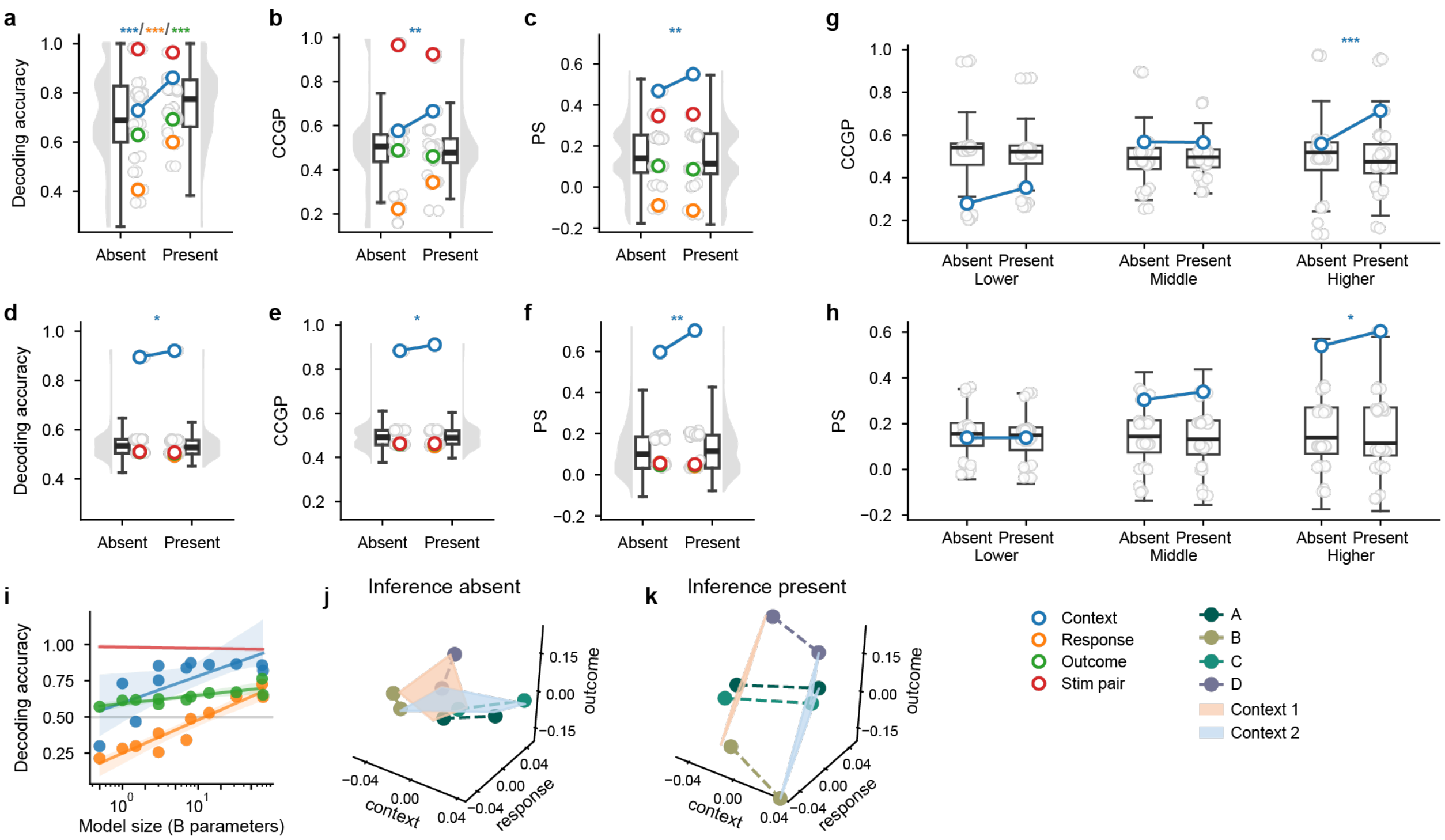

**Fig. 2 | Human-like abstract geometry selectively emerges in large language models during inference. a–c**, Stimulus-stage representations, measured after the current stimulus has been inserted and immediately before response generation. Raincloud plots compare inference absent and inference present sessions for decoding accuracy (a), cross-condition generalization performance (CCGP) (b), and parallelism score (PS) (c). Half-violins and box plots summarize the distribution of values across the 35 balanced dichotomies; boxes show the median and interquartile range, and whiskers extend to 1.5× the interquartile range. Each point corresponds to one balanced dichotomy. Grey points show the unnamed dichotomies, whereas colored open circles mark the named dichotomies—context, response, outcome and stimulus pair—with connecting lines indicating their paired inference absent to inference present changes. **d**–**f**, Baseline-stage representations, measured before insertion of the current stimulus, shown using the same conventions as in a–c. **g**,**h**, Depth-block analysis of stimulus-stage geometry. CCGP (g) and PS (h) are shown across lower, middle and higher transformer blocks, using the same inference absent versus inference present comparison. **i**, Scaling of representational geometry across model size. For each model, values were averaged across evaluation sessions to give one point per model; the x axis indicates parameter count in billions on a log scale. Solid lines show linear fits against log10 (parameter count), and shaded bands denote bootstrap 95% confidence intervals. **j**,**k**, Representative three-dimensional projections of task-state geometry in inference absent (j) and inference present (k) sessions. Colors indicate stimulus identity, and translucent surfaces connect conditions sharing the same context. These projections are used for visualization only; projection procedures and statistical analyses in the full representational space are described in Methods. Asterisks in a–h mark significant effects for named dichotomies only. Asterisks are assigned when three criteria are jointly satisfied: the inference present mean exceeds the inference absent mean, the inference present value exceeds the 95th percentile of the empirical null distribution, and the session-level permutation test is significant after Benjamini–Hochberg correction across 35 dichotomies. *$q < 0.05$, **$q < 0.01$, ***$q < 0.001$. Details of null construction, resampling and statistical testing are provided in Methods.

## A higher-layer abstract context geometry marks inference in language models

We used the hippocampal findings of Courellis et al. as a neural benchmark for interpreting LLM representational geometry[1]. In that study, inference was associated with abstract population geometry in the human hippocampus: during stimulus processing, context and stimulus-pair structure were represented in an abstract format, whereas during the baseline period, abstract coding was more selectively expressed as context geometry in inference-present sessions. We therefore asked whether LLM hidden states exhibit a related geometric signature of inference, and whether its stage profile reveals model-specific mechanisms.

We first examined stimulus-stage representations across all balanced dichotomies, including the named variables context, response, outcome and stimulus pair, to determine whether inference was associated with abstract task geometry rather than a nonspecific increase in linear readability (Fig. 2a–c). Decoding accuracy improved for several task variables, including context, response and outcome, indicating that task-relevant information became more linearly readable after stimulus integration. However, increased readability did not imply a global increase in abstraction. Among the named variables, the coordinated enhancement of abstract geometric metrics was most evident for latent context: context CCGP increased by $\Delta = +0.090$ (CI95 [+0.049, +0.130], $q = 5.25 \times 10^{-3}$), and context PS increased by $\Delta = +0.081$ (CI95 [+0.048, +0.114], $q = 8.75 \times 10^{-3}$) (Fig. 2b,c). Stimulus pair remained at a relatively high level across both inference-absent and inference-present sessions, indicating that stimulus-related structure was already stably represented at the stimulus stage. Thus, inference was associated with a selective strengthening of abstract task geometry, rather than a nonspecific increase in representational readability.

We next examined whether this inference-related geometry was already present at the baseline stage, before the current stimulus was introduced (Fig. 2d–f). This analysis revealed a stage profile that was related to, but distinct from, the hippocampal benchmark. In LLMs, the baseline-stage profile was related but distinct: latent-context geometry was already prominent before the current stimulus was introduced, with context decoding accuracy and CCGP both near 0.90 even in inference-absent sessions and increasing further to approximately 0.93 in inference-present sessions (decoding accuracy: $\Delta = +0.026$, CI95 [+0.015, +0.036], $q = 1.93 \times 10^{-2}$; CCGP: $\Delta = +0.027$, CI95 [+0.016, +0.039], $q = 4.2 \times 10^{-2}$). Context PS showed the most pronounced inference-related change, increasing from approximately 0.60 to 0.70 ($\Delta = +0.104$, CI95 [+0.075, +0.135], $q = 3.5 \times 10^{-3}$). Thus, the baseline-stage pattern suggests that LLMs maintain a strong carried-over task-state representation before the current stimulus is processed: context CCGP and PS were higher at baseline than at the stimulus stage, and inference-present sessions further increased the regularity of this already available geometry, most clearly through stronger directional consistency. A plausible explanation is that the autoregressive text history explicitly preserves prior stimulus–response–outcome evidence, allowing latent context to be represented before stimulus integration, whereas successful inference depends on organizing that carried-over state into a more stable and generalizable context axis.

Having identified inference-related abstract context geometry, we next asked whether it was uniformly distributed across model depth or preferentially enriched in specific transformer layers.

To probe this, we divided model depth into three contiguous depth ranges corresponding to lower, middle and higher layers, and computed decoding accuracy, CCGP and PS within each range (Extended Data Fig. ___ and Fig. 2g,h). Context showed the strongest simultaneous enhancement in the higher layers: decoding accuracy increased by $\Delta = +0.180$ (CI95 [+0.146, +0.215], $q = 1.75 \times 10^{-4}$), CCGP by $\Delta = +0.155$ (CI95 [+0.113, +0.194], $q = 2.19 \times 10^{-4}$), and PS by $\Delta = +0.064$ (CI95 [+0.028, +0.101], $q = 3.68 \times 10^{-2}$). Lower and middle layers showed weaker or inconsistent effects. This depth specificity indicates that abstract context geometry is not uniformly distributed across the model, but is preferentially enriched in a higher-layer functional band. This higher-layer enrichment supports a computational analogy to the hippocampal population geometry reported in humans: in LLMs, higher transformer layers appear to provide a functional locus for organizing and applying latent task context during inference.

Across models, context geometry tended to strengthen with model scale (Fig. 2i and Extended Data Fig. ___). Context decoding accuracy showed a significant positive correlation with model size, and both CCGP and PS showed positive trends, whereas response exhibited a dissociation between increased decoding accuracy and reduced directional consistency. Finally, three-dimensional visualization provided an intuitive summary of the same geometric transformation (Fig. 2j,k). In inference absent sessions, the eight condition centroids were more entangled, whereas inference present sessions showed clearer separation between context planes and more consistent cross-context displacement. Together, these results show that inference in language models is marked by a higher-layer-enriched abstract context geometry: context information is already available in the accumulated task state, but successful inference is associated with its more regular, cross-condition-generalizable organization.

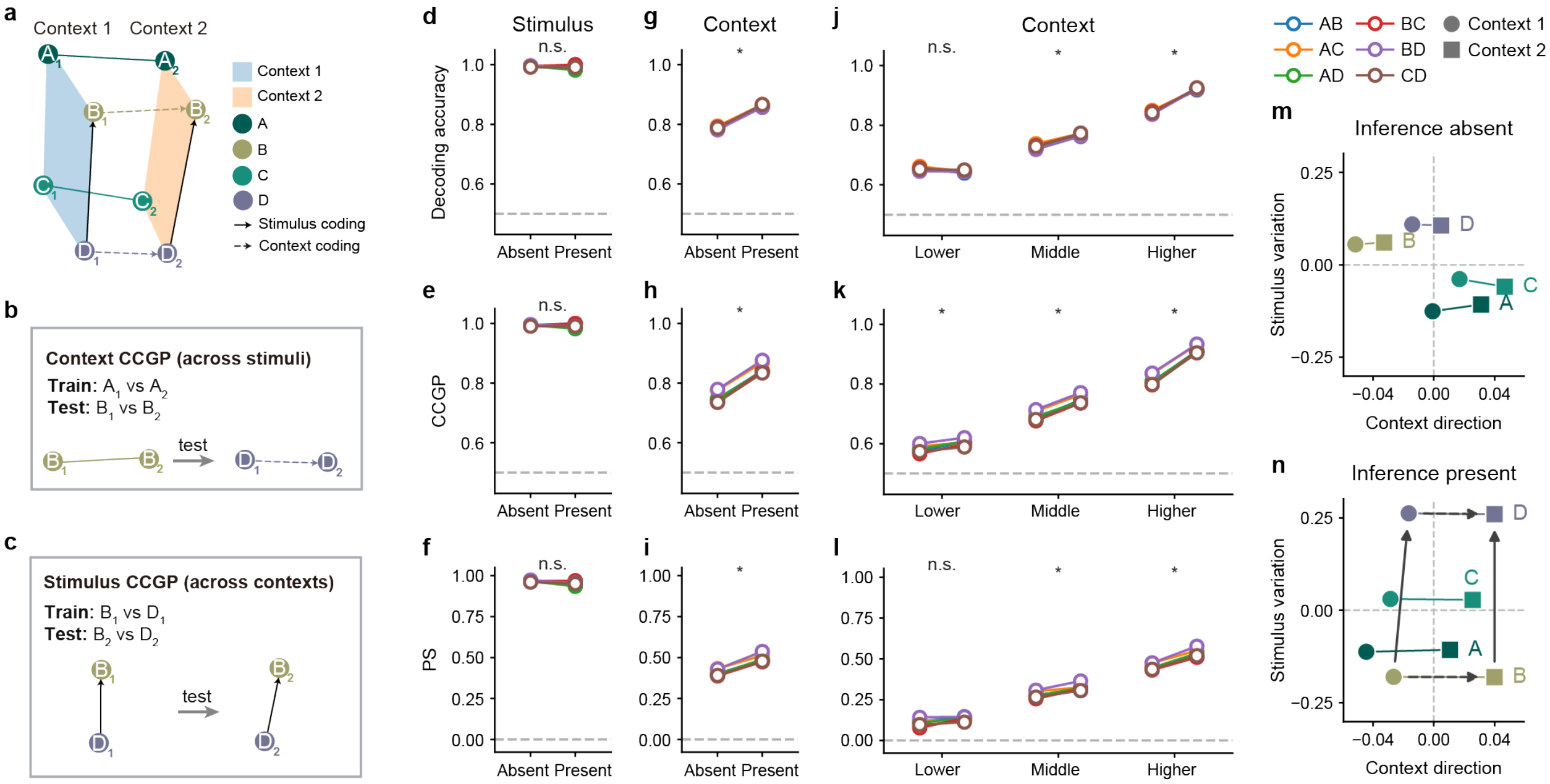

**Fig. 3 | Latent context is represented as a stimulus-invariant axis in higher transformer layers during inference. a**, Schematic of the stimulus–context geometry. Each point denotes the representation of one stimulus under one latent context. Solid arrows indicate stimulus coding directions within a context, and dashed arrows indicate context coding directions connecting the same stimulus across contexts. A factorized stimulus–context

geometry predicts that context coding directions are approximately parallel across stimuli, while stimulus coding directions are preserved across contexts. **b**, Context CCGP across stimuli. A linear decoder is trained to discriminate the two contexts using one stimulus pair and tested on a held-out stimulus pair. This analysis asks whether the context code generalizes across stimulus identity. **c**, Stimulus CCGP across contexts. A linear decoder is trained to discriminate two stimuli in one context and tested on the same stimulus pair in the other context. This analysis asks whether stimulus identity is represented consistently across latent contexts. **d**–**f**, Stimulus representations across contexts. Mean decoding accuracy (**d**), cross-condition generalization performance (CCGP; **e**) and parallelism score (PS; **f**) are shown for all six stimulus pairs. Open circles denote individual stimulus-pair contrasts, and lines connect the same pair between inference absent and inference present sessions. Stimulus identity is strongly represented in both groups, but neither stimulus CCGP nor stimulus PS shows a significant inference-related increase. **g**–**i**, Context representations across stimulus pairs. Mean decoding accuracy (**g**), CCGP (**h**) and PS (**i**) are shown using the same plotting convention as in **d**–**f**, but now for discriminating latent context across different stimulus pairs. Inference present sessions show higher context decoding, context CCGP and context PS, indicating that successful inference is associated with a context code that is not only more linearly readable, but also more consistent and reusable across stimulus identities. **j**–**l**, Depth-resolved context geometry across lower, middle and higher transformer layers. Context decoding accuracy (**j**), CCGP (**k**) and PS (**l**) are computed separately within each depth range. Context representations become progressively stronger and more geometrically regular across depth, with the highest values in higher layers, indicating that the abstract context axis is preferentially expressed in late transformer representations. **m**,**n**, Low-dimensional visualization of the stimulus–context geometry in inference absent (**m**) and inference present (**n**) sessions. Condition centroids are projected onto a context direction and an orthogonal stimulus-variation axis. Colors denote stimulus identities and marker shapes denote latent contexts. In inference absent sessions, cross-context displacements vary across stimuli, indicating an entangled stimulus–context geometry. In inference present sessions, corresponding stimuli are displaced in a more parallel direction across contexts, consistent with the emergence of a stimulus-invariant context axis. Asterisks indicate permutation-test significance (*$P < 0.05$); n.s., not significant. For **d**–**i**, significance is assessed for the pair-averaged inference-present versus inference-absent difference across the six stimulus-pair contrasts after averaging across sessions. For **j**–**l**, the same test is applied separately within each depth range. Asterisks marked comparisons additionally require the pair-averaged inference-present value to exceed both the inference-absent value and the 95th percentile of the inference-present null distribution. Full null construction, resampling procedure and statistical tests are described in Methods.

### Inference disentangles stimulus identity and latent context in higher transformer layers

We next focused on stimulus-stage representations to ask whether the latent context code generalizes across arbitrary pseudoword stimuli. This generalization is the central computational demand of the task: after a context switch, feedback from one stimulus should guide responses to other stimuli that have not yet appeared under the new context. Because the pseudowords were arbitrary, any such transfer should depend on the organization of context relative to stimulus identity rather than on semantic structure among the stimuli. We therefore reorganized the eight task states as stimulus × context and used pairwise CCGP and PS analyses to test whether context coding generalized across stimuli and whether stimulus coding was preserved across contexts (Fig. 3a–c).

In this stimulus–context geometry, stimulus identity was strongly decodable and highly cross-context generalizable in both inference-absent and inference-present sessions (Fig. 3d–f). Across the six stimulus-pair contrasts, stimulus decoding, CCGP and PS were all close to ceiling and showed no significant inference-related increase (decoding, 0.992 versus 0.994; CCGP, 0.992 versus 0.994; PS, 0.954 versus 0.966; all n.s.). Thus, the model maintained a stable representation of stimulus identity regardless of whether inference was expressed. The inference-dependent change instead appeared in the organization of latent context across stimulus pairs (Fig. 3g–i). Context decoding, CCGP and PS were all higher in inference-present than inference-absent sessions (decoding, 0.862 versus 0.788; CCGP, 0.850 versus 0.754; PS, 0.495 versus 0.407; all $P < 0.05$). These results indicate that successful inference was associated with a more disentangled stimulus–context geometry, in which latent context was separated from stimulus identity and could be applied across arbitrary pseudoword stimuli.

Layer-block analysis further separated the depth profiles of stimulus and context geometry (Fig. 3j–l and Extended Data Fig. __). Stimulus geometry was already well established in lower layers and remained comparatively stable across middle and higher layers. By contrast, context abstraction increased progressively with depth and reached its highest values where context decoding, CCGP and PS were all higher in inference-present sessions (decoding, 0.843 to 0.922; CCGP, 0.814 to 0.916; PS, 0.451 to 0.534; all $P < 0.05$). This depth profile provides further evidence for stimulus–context disentanglement: stable stimulus geometry is preserved across depth, while higher layers progressively organize a shared context axis over these stimulus states.

The stimulus–context reorganization described above was also visible in an all-layer low-dimensional projection of the condition centroids (Fig. 3m,n), providing a summary visualization of the geometry across the model. In inference-absent sessions, the two contexts did not form a consistent relationship across stimuli, indicating that stimulus identity and latent context remained entangled. In inference-present sessions, corresponding stimuli across contexts were related by a more parallel context direction, consistent with a shared context axis. This geometry provides an intuitive account of the CCGP and PS results: the inferred context state can be applied across different stimuli, allowing feedback from one stimulus to guide responses to others. Together, these results show that successful inference is accompanied by a selective reorganization of stimulus–context geometry. Stimulus identity remains strongly represented, whereas latent context becomes increasingly stimulus-invariant and concentrated in higher transformer layers, providing a representational substrate for generalizable rule updating.

**Task-sequence language modelling induces abstract context geometry**

Having shown that inference-present sessions are marked by abstract geometry, we next asked whether learning the sequence structure of the task can induce both generalizable inference and the corresponding representational organization. To this end, we fine-tuned base models on independent reversal-learning task sequences, varying stimuli and trial orders, using only a language-modelling objective. We then compared base and finetuned models under the same evaluation protocol. Behavioral results showed that language modelling fine-tuning significantly improved the prevalence of generalizable inference in the task: the proportion of inference-present sessions in finetuned models was markedly higher than in base models (Fig. 4a, 76.1% vs 16.2%), indicating that models more frequently entered a context-level inference regime rather than relying on local stimulus–response matching. Furthermore, in the trajectory of inference trials after a context switch, both groups showed a performance drop far below chance on the crucial first trial after the switch but finetuned models recovered faster and reached higher accuracy in the subsequent reasoning window (Fig. 4b, base: 0.45±0.01, finetuned improved to 0.77±0.01). These behavioral changes indicate that task-sequence language modelling increased the model's ability to update latent-context hypotheses from sparse feedback, beyond improving local trial-level prediction alone.

This behavioral improvement was accompanied by a systematic reorganization of context geometry. Context decoding accuracy remained essentially unchanged in the lower block (Δ≈0; Fig. 4c) but rose significantly in middle and higher blocks, manifesting as a robust improvement when aggregated across all layers (base/finetuned: 0.750→0.869, Δ=+0.119, CI95 [0.095, 0.144], $q=1.09\times10^{-4}$; Fig. 4c). We further observed that CCGP and PS also increased, with stronger effects in middle and higher layers than in lower layers. Across all layers, CCGP increased from 0.591 to 0.666 (Δ=+0.075, CI95 [0.048, 0.102], $q=1.59\times10^{-4}$; Fig. 4d), and PS from 0.481 to 0.638 (Δ=+0.157, CI95 [0.134, 0.179], $q=1.25\times10^{-4}$; Fig. 4e). These results indicate this task did more than improve context readout: it made the context code more cross-condition generalizable and directionally consistent. The side-by-side geometries in Fig. 4f,g provide a compact visualization of this change , showing a clearer context axis and more regular cross-context correspondence in the finetuned model, in line with the CCGP and PS increases. In summary, task sequence language modelling induces abstract context geometry without explicit geometric supervision,  providing a link from improved inference to a more abstract representation.

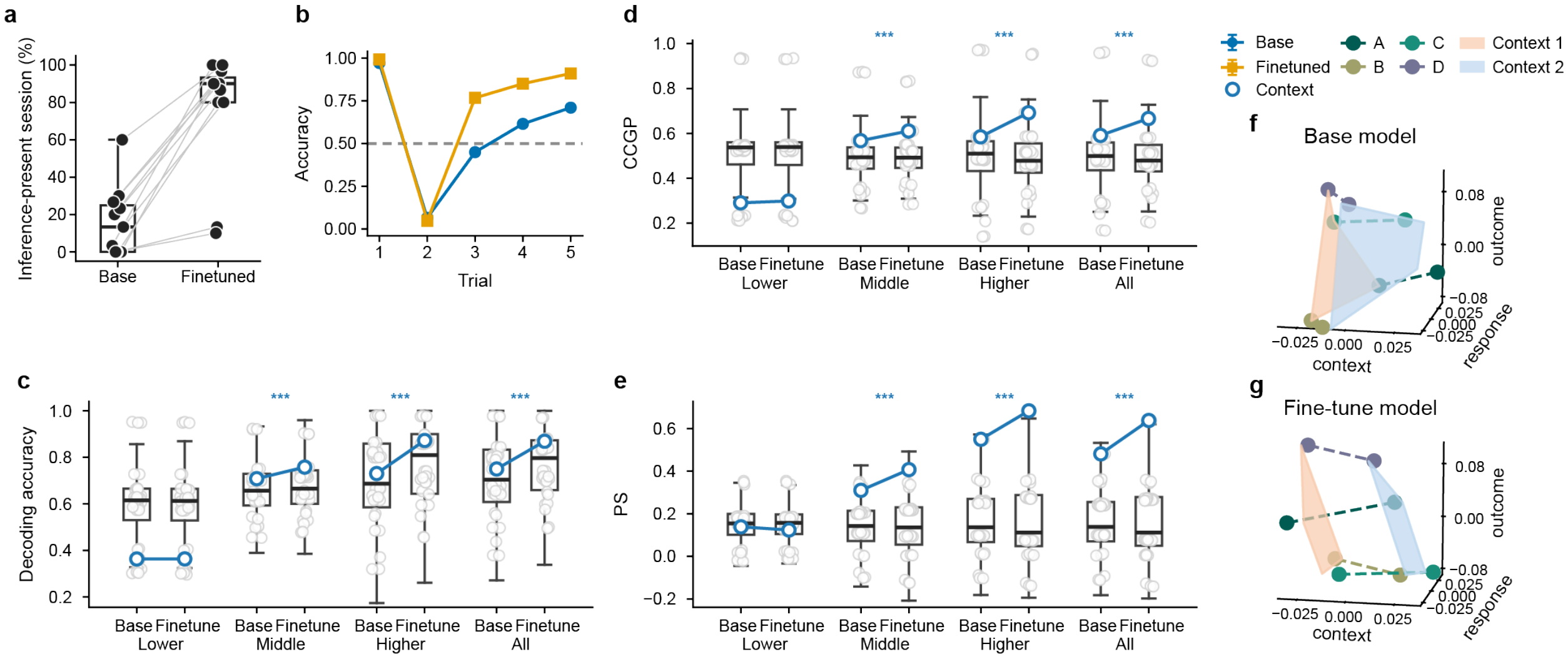

**Fig. 4 | Task-sequence language modelling induces the emergence of abstract context representations**. **a**, Proportions of inference-present sessions for the same set of base models before (Base) and after task-sequence language-model (LM) finetuning (Finetuned). **b**, Accuracy around the context switch. **c–e**, Block-wise representational geometry for the context dichotomy comparing Base and Finetuned models across lower, middle, higher blocks and across all layers: decoding accuracy (c), cross-condition generalization performance (CCGP) (d), and parallelism score (PS) (e). Grey dots indicate individual dichotomies; boxplots show median (center line), interquartile range (box), and 1.5×IQR whiskers. Blue open circles connected by lines denote the context dichotomy mean for each session and highlight the Base→Finetuned change within each depth block. Stars indicate significant Base–Finetuned differences (two-sided session-level permutation test adjusted for 35 dichotomies using the Benjamini–Hochberg procedure (FDR); *** $q < 0.001$). **f,g**, Representative geometry visualization of task-state representations projected into the three-dimensional subspace spanned by linear decoding axes for context, response, and outcome. Points correspond to mean representations for stimuli A–D under each context; translucent planes depict

the within-context affine structure, and dashed segments connect corresponding stimulus states across contexts, illustrating a more consistent context translation in the finetuned model.

## Geometric constraints enhance generalizable inference

We next tested the complementary direction: whether directly constraining representational geometry, without a task-sequence language-modelling objective, can improve generalizable inference. We introduced a geometric-structure constraint (GeoStruct) on the 2×2×2 context–response–outcome condition cube. For each latent factor, we computed difference vectors across the four settings of the remaining variables and minimized the cosine distance between these vectors. This objective encourages coding directions to remain consistent across task backgrounds, aligning the representation with the directional parallelism and cross-condition generalization measured by PS and CCGP. At the behavioral level, GeoStruct increased the prevalence of generalizable inference, raising the proportion of inference-present sessions from 16.2% in base models to 43.2% after geometric-constraint training (Fig. 5a). It also systematically improved inference quality within the reasoning window (Fig. 5b). On the first inference trial, accuracy for GeoStruct improved from the base of 0.45±0.01 to 0.63±0.01 (Δ=+0.18), and the subsequent recovery plateau also shifted upward overall (e.g., the endpoint improved from 0.71±0.01 to 0.82±0.01; Δ=+0.12). These behavioral gains indicate that a geometry-focused intervention can increase both the frequency and quality of generalizable inference, providing interventional evidence that abstract representational structure is functionally relevant to reasoning.

To localize where in the model this geometric intervention was most effective, we apply GeoStruct to lower, middle, higher or all layers and measuring both behavioral gains and context geometry (Fig. 5c–f). When supervision was applied only to the lower layers, Behavioral improvement was weakest (Δaccuracy=+0.048, CI95 [+0.015, +0.083], Bonferroni-corrected P=0.100, m=4 comparisons; Fig. 5c–d). Correspondingly, lower-layer supervision produced only modest geometric changes: CCGP/PS at the lower probe increased from 0.29/0.14 to 0.36/0.28, whereas middle and higher probes showed no systematic improvement (CCGP ~0.56–0.60, PS ~0.36–0.56; Fig. 5e–f). This suggests that constraining lower-layer geometry alone was insufficient to strengthen the middle-to-higher-layer abstract context geometry associated with robust behavioral gains. By contrast, constraining middle or higher layers produced much larger behavioral gains (middle: Δ=+0.166, CI95 [+0.118, +0.223], P=0.0052; higher: Δ=+0.206, CI95 [+0.146, +0.272], P=0.0052; Fig. 5c–d). These gains were accompanied by stronger middle-to-higher-layer geometry: middle supervision raised middle/higher-probe CCGP to ~0.66–0.67 and PS to ~0.67–0.78, whereas higher supervision further increased higher-probe PS to ~0.87 (Fig. 5e–f). Applying GeoStruct to all layers also improved behavior (Δ=+0.178, CI95 [+0.120, +0.249], P=0.0052) and produced geometric enhancement mainly in middle and higher probes. Overall, the effect was concentrated in the middle-to-higher layers, matching the depth profile of abstract context geometry identified above.

Finally, geometric visualization provided intuitive support for the quantitative conclusions above (Fig. 5g,h). Compared to the base model's more entangled organization, the GeoStruct model appeared closer to a structure with more consistent cross-context correspondence. Together with the task-sequence language modelling results, GeoStruct provides the complementary direction of interventional evidence: strengthening abstract geometry can improve generalizable inference even without language-model supervision. These two interventions support a functional link between abstract representational geometry and reasoning in LLMs.

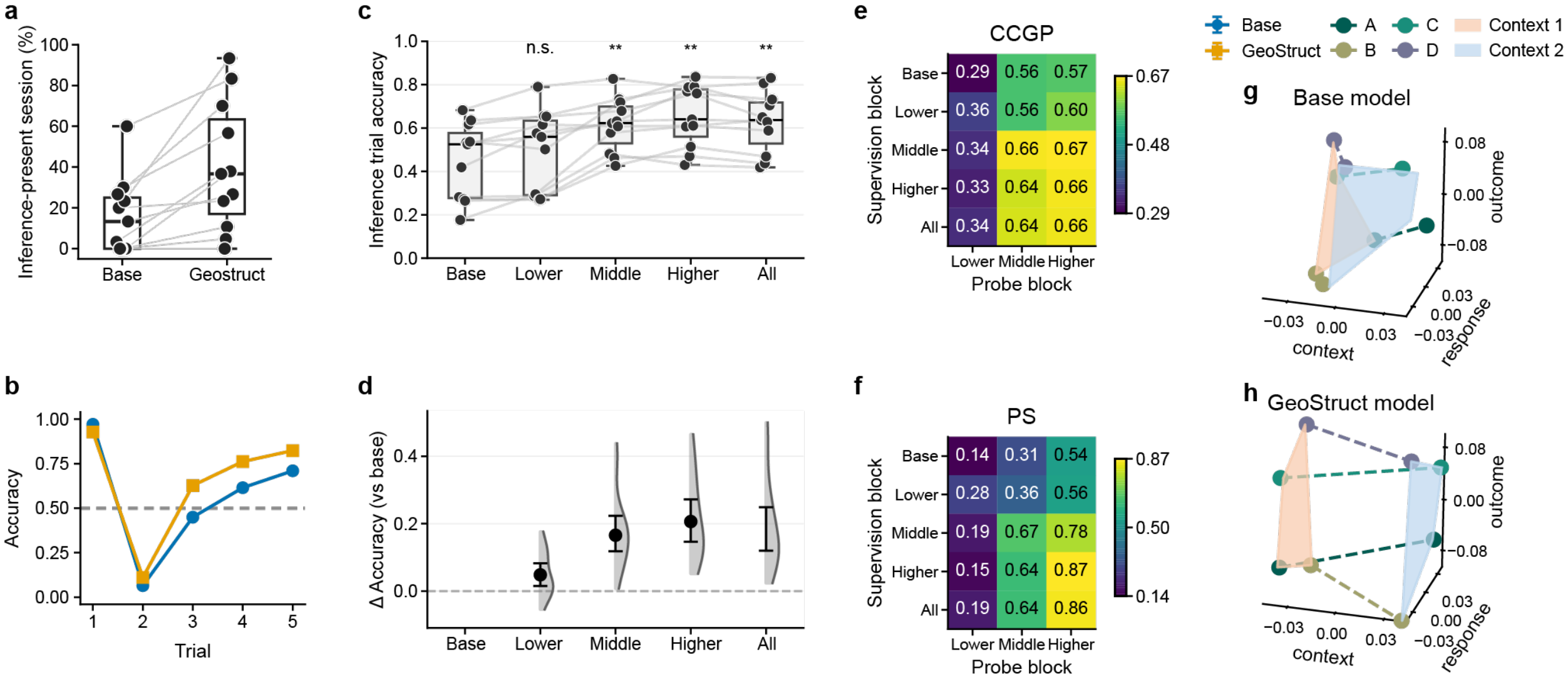

**Fig. 5 | Geometric constraints enhance inference performance and reshape abstract geometry without an language modelling objective**. **a**, Proportions of inference-present sessions for the base model and the geometry-constrained model (GeoStruct), trained using only representational geometric-constraint loss (no task-sequence language-modeling objective). **b**, Trial-by-trial accuracy around the inference window for base and GeoStruct models. The x-axis indexes key trials (Last, 1–4); the dashed line indicates chance level. Points show the mean across sessions and error bars denote s.e.m. The mean ± s.e.m. at the first inference trial is annotated. **c**, Accuracy on the first inference trial when GeoStruct supervision is restricted to different depth blocks (lower, middle, higher, or all). Each dot denotes one model's mean accuracy (averaged across evaluation sessions); grey lines connect paired model variants derived from the same base model across supervision conditions. Boxplots show the median and interquartile range (IQR); whiskers extend to 1.5×IQR. Asterisks indicate significance for paired contrasts versus Base after Bonferroni correction across four comparisons (*P<0.05, **P<0.01, ***P<0.001; n.s., not significant; paired sign-flip permutation test, 10,000 permutations). **d**, Accuracy gain relative to Base (Δaccuracy vs. Base) for each supervision block. Black dots indicate the mean Δ across models; error bars show bootstrap 95% confidence intervals (10,000 resamples); grey violins depict the distribution of model-wise Δ values. **e-f**, Mechanistic analysis: how the supervision block (rows) reshapes abstraction measured at different probe depths (columns) using CCGP (e) and PS (f). Each cell shows the mean across models; color encodes metric magnitude. Both heatmaps exhibit a pronounced block structure, with selective increase in the middle/higher probe submatrix under middle/higher (and all) supervision, consistent with preferential strengthening of an abstract context axis at intermediate-to-deep depths with limited transfer to adjacent depths. **g**, Geometric visualization of the 2×2×2 condition cube for the base and GeoStruct models. Points corresponding to four stimuli (A–D) form two context-specific planes (shaded). Dashed lines connect the same stimulus across contexts. The geometry-constrained model exhibits a more consistent cross-context translation/parallel structure, consistent with increased PS/CCGP and improved cross-condition generalization.

## Discussion

This study identifies abstract representational geometry as a mechanistically informative framework for analysing inference in large language models. In a text-based reversal-learning task, a subset of LLM sessions exhibited rapid context-level updating rather than gradual stimulus-by-stimulus relearning. These inference-present sessions were associated with a more abstract latent-context geometry: context coding showed greater cross-condition generalizability, stronger directional consistency and clearer separation from stimulus identity. This geometry was not uniformly distributed across model depth, but was preferentially expressed in middle-to-higher transformer layers. Together with the behavioural and interventional results, these findings support the central claim that inference in LLMs is shaped by abstract representational geometry at the neural-population level, where latent task structure is organized into a reusable, stimulus-generalizing context code.

Human hippocampal recordings therefore serve here as a comparative framework for interpreting LLM representational geometry. Courellis et al. provided a neural benchmark showing that human hippocampal populations form abstract task-variable geometry during inference[1]. By adapting the same behavioural logic and related geometric measures to LLM hidden states, we observed a corresponding geometric signature: successful inference was accompanied by a more regular context geometry that generalized across task conditions and stimulus identities. At the same time, the LLM stage profile differed from the hippocampal benchmark. In particular, baseline-stage context geometry was already prominent before the current stimulus was introduced, consistent with the autoregressive text history maintaining a carried-over latent task state. Thus, the comparison points to a shared computational motif: abstract variables can support generalization when encoded by reusable and geometrically consistent coding directions.

The depth profile of this geometry suggests a functional organization within transformer representations. Stimulus identity was already robustly represented in lower layers and remained comparatively stable across depth, whereas latent context became more abstract, more stimulus-invariant and more directionally consistent in middle-to-higher layers. This dissociation points to a division of representational roles: the model maintains stimulus information while organizing a latent context state that can be applied across stimuli. Together, these results indicate that higher transformer layers provide a functional locus in which abstract context geometry is organized for generalizable rule updating.

The two intervention experiments further support a functional link between geometry and inference. Task-sequence language modelling increased the prevalence of generalizable inference and induced a more abstract context geometry without explicit geometric supervision. Conversely, GeoStruct constrained coding-direction consistency and improved inference without a language-modelling objective. Together, these interventions indicate that abstract geometry is functionally involved in reasoning, extending interpretability beyond feature-level descriptions. Sparse autoencoders and linear probes can identify semantic or conceptual directions, whereas the present analysis asks whether task variables are arranged so that linear readouts can generalize across conditions. Representational geometry therefore links local feature attribution to distributed, population-level reasoning dynamics.

suggests that abstract geometry can serve as a methodological tool for investigating shared computational principles between biological and artificial intelligence, by revealing how different systems organize latent task variables during inference. Most critically, this causal chain provides a theoretical basis for "Mechanistic Alignment"—ensuring that models truly internalize human logical rules rather than learning shortcut strategies by real-time monitoring of the decoupling degree of neural manifolds, rather than relying solely on easily compromised Behavioural fine-tuning (e.g., RLHF). By providing a transparent, internal metric for reasoning fidelity, our geometric framework contributes a crucial verification tool for the safe development of Artificial General Intelligence (AGI), directly addressing the urgent call for robust oversight mechanisms to manage extreme AI risks amid rapid progress[43]. Although LLMs express a hippocampal-like geometric motif, their inference prevalence remains below that of humans, indicating that current transformer architectures still fall short of the robustness of biological inference under sparse feedback. Future work should test whether richer representational structures, including curved manifolds, attractor dynamics or non-Euclidean task geometries[44–46], can improve the reliability and prevalence of generalizable inference in artificial systems.

In conclusion, this study identifies abstract representational geometry as a shared analytical framework for comparing inference in humans and LLMs. By revealing this deep isomorphism, we not only decipher the internal operating mechanisms of "black box" models from a geometric perspective but also lay a mechanistic foundation for building safe, trustworthy, and human-like reasoning AGI in the future.

## References

1. Courellis, H. S. *et al.* Abstract representations emerge in human hippocampal neurons during inference. *Nature* **632**, 1104–1111 (2024).
2. Behrens, T. E. *et al.* What is a cognitive map? Organizing knowledge for flexible Behaviour. *Neuron* **100**, 490–509 (2018).
3. Tenenbaum, J. B., Kemp, C., Griffiths, T. L. & Goodman, N. D. How to grow a mind: Statistics, structure, and abstraction. *Science* **331**, 1279–1285 (2011).
4. Summerfield, C., Luyckx, F. & Sheahan, H. Structure learning and the posterior parietal cortex. *Progress in Neurobiology* **184**, 101717 (2020).
5. O'Keefe, J. & Nadel, L. *The Hippocampus as a Cognitive Map*. (Clarendon Press, 1978).
6. Eichenbaum, H. On the functional organization of the hippocampal memory system. *Proceedings of the National Academy of Sciences* **114**, 10565–10567 (2017).
7. Bellmund, J. L., Gärdenfors, P., Moser, E. I. & Doeller, C. F. Navigating cognition: Spatial codes for human thinking. *Science* **362**, eaat6766 (2018).
8. Whittington, J. C. *et al.* The Tolman-Eichenbaum Machine: Unifying space and relational memory through generalization in the hippocampal formation. *Cell* **183**, 1249–1263 (2020).
9. Bernardi, S. *et al.* The geometry of abstraction in the hippocampus and prefrontal cortex. *Cell* **183**, 954–967 (2020).
10. Constantinescu, A. O., O'Reilly, J. X. & Behrens, T. E. Organizing conceptual knowledge in humans with a grid-like code. *Science* **352**, 1464–1468 (2016).
11. Fusi, S., Miller, E. K. & Rigotti, M. Why neurons mix: high dimensionality for higher cognition. *Current Opinion in Neurobiology* **37**, 66–74 (2016).
12. Higgins, I. *et al.* beta-VAE: Learning Basic Visual Concepts with a Constrained Variational Framework. in *International Conference on Learning Representations* (2017).
13. Eastwood, C. & Williams, C. K. A Framework for the Quantitative Evaluation of Disentangled Representations. in *International Conference on Learning Representations* (2018).
14. Lake, B. M., Ullman, T. D., Tenenbaum, J. B. & Gershman, S. J. Building machines that learn and think like people. *Behavioural and Brain Sciences* **40**, e253 (2017).

15. Badre, D., Doll, B. B., Long, N. M. & Frank, M. J. Rostrolateral prefrontal cortex and individual differences in uncertainty-driven exploration. *Neuron* **73**, 595–607 (2012).
16. Richards, B. A. *et al.* A deep learning framework for neuroscience. *Nat Neurosci* **22**, 1761–1770 (2019).
17. Sorscher, B., Mel, G. C., Ocko, S. A., Giocomo, L. M. & Ganguli, S. A unified theory for the computational and mechanistic origins of grid cells. *Neuron* **111**, 121-137.e13 (2023).
18. Sterling, P. & Laughlin, S. *Principles of Neural Design*. (The MIT Press, Cambridge, Massachusetts, 2015).
19. Webb, T., Holyoak, K. J. & Lu, H. Emergent analogical reasoning in large language models. *Nat Hum Behav* **7**, 1526–1541 (2023).
20. Brown, A., Smith, J. & al, et. The Emergence of Logical Consistency in Reasoning Models. *Nature Machine Intelligence* **7**, 112–128 (2025).
21. Garrido, Q. & al, et. Beyond Text: Do LLMs Build Implicit World Models during Pre-training? in *International Conference on Learning Representations (ICLR)* (2025).
22. OpenAI. *GPT-5 Technical Report: Reasoning, Abstraction, and Generalization*. https://openai.com/index/gpt-5-technical-report/ (2025).
23. Huang, J. & Chang, K. C.-C. Towards Reasoning in Large Language Models: A Survey. in *Findings of the Association for Computational Linguistics: ACL 2023* (eds Rogers, A., Boyd-Graber, J. & Okazaki, N.) 1049–1065 (Association for Computational Linguistics, Toronto, Canada, 2023). doi:10.18653/v1/2023.findings-acl.67.
24. Geirhos, R. *et al.* Shortcut learning in deep neural networks. *Nature Machine Intelligence* **2**, 665–673 (2020).
25. McCoy, R. T., Pavlick, E. & Linzen, T. Right for the Wrong Reasons: Diagnosing Syntactic Heuristics in Natural Language Inference. in *Proceedings of the 57th Annual Meeting of the Association for Computational Linguistics* 3428–3448 (2019).
26. Bricken, T. *et al.* Towards Monosemanticity: Decomposing Language Models with Dictionary Learning. *Transformer Circuits Thread* https://transformer-circuits.pub/2023/monosemantic-features (2023).
27. Cunningham, H., Ewart, A., Riggs, L., Huben, R. & Sharkey, L. Sparse Autoencoders Find Interpretable Features in Large Language Models. in *International Conference on Learning Representations* (2024).
28. Park, K., Choe, Y. J. & Veitch, V. The Linear Representation Hypothesis and the Geometry of Large Language Models. *arXiv preprint arXiv:2311.03658* (2023).

29. Mikolov, T., Chen, K., Corrado, G. & Dean, J. Efficient Estimation of Word Representations in Vector Space. in *International Conference on Learning Representations* (2013).

30. Burns, C., Ye, H., Klein, D. & Steinhardt, J. Discovering Latent Knowledge in Language Models Without Supervision. in *International Conference on Learning Representations* (2023).

31. Saxena, S. & Cunningham, J. P. Towards the neural population doctrine. *Current Opinion in Neurobiology* **55**, 103–111 (2019).

32. Rigotti, M. *et al.* The importance of mixed selectivity in complex cognitive tasks. *Nature* **497**, 585–590 (2013).

33. Mante, V. Context-dependent computation by recurrent dynamics in prefrontal cortex. *Nature* **503**, 78–84 (2013).

34. Libby, A. & Buschman, T. J. Rotational dynamics reduce interference between sensory and memory representations. *Nature Neuroscience* **24**, 715–726 (2021).

35. Peel, A., Vickers, C. & Johnston, V. Contextual reversal learning relies on hippocampal-prefrontal interactions. *Scientific Reports* **12**, 4512 (2022).

36. Dubey, A., Jauhri, A., Pandey, A., Keshwam, A., & others. The Llama 3 Herd of Models. *arXiv preprint arXiv:2407.21783* (2024).

37. Yang, A., Yang, B., Hui, B., Zheng, B., & others. Qwen2 Technical Report. *arXiv preprint arXiv:2407.10671* (2024).

38. Anthropic. *Claude 4.5 Model Card*. https://www.anthropic.com/research/claude-4-5-sonnet (2025).

39. Caucheteux, C. & King, J.-R. Brains and algorithms partially converge in natural language processing. *Communications Biology* **5**, 1–10 (2022).

40. Schrimpf, M. *et al.* The neural architecture of language: Integrative modeling converges on predictive processing. *Proceedings of the National Academy of Sciences* **118**, e2105646118 (2021).

41. Ouyang, L. *et al.* Training language models to follow instructions with human feedback. in *Advances in Neural Information Processing Systems* vol. 35 27730–27744 (2022).

42. Kaplan, J. *et al.* Scaling Laws for Neural Language Models. *arXiv:2001.08361 [cs, stat]* http://arxiv.org/abs/2001.08361 (2020).

43. Managing extreme AI risks amid rapid progress. https://www.science.org/doi/10.1126/science.adn0117 doi:10.1126/science.adn0117.

44. Khona, M. & Fiete, I. R. Attractor and integrator networks in the brain. *Nature Reviews Neuroscience* **23**, 744–766 (2022).

45. Nickel, M. & Kiela, D. Poincaré Embeddings for Learning Hierarchical Representations. in *Advances in Neural Information Processing Systems* vol. 30 (2017).

46. Bronstein, M. M., Bruna, J., LeCun, Y., Szlam, A. & Vandergheynst, P. Geometric deep learning: going beyond Euclidean data. *IEEE Signal Processing Magazine* **34**, 18–42 (2017).

# Methods

## Task and session generation

We used an implicit context-switch reversal-learning task with two latent contexts and four stimuli. To minimize semantic priors and ensure cross-session comparability, stimuli were displayed as pseudo-words that were re-sampled and re-assigned for each session. Specifically, for every session we randomly drew four pseudo-words from a shared candidate pool (e.g., “baf”, “teq”, “zir”) and mapped them to the four stimulus identities (1–4), yielding session-specific visible stimulus labels.

Each context defined a fixed stimulus–response–outcome (S–R–O) mapping over the four stimuli, with responses restricted to {left, right}. On each trial, the stimulus was sampled from the four S–R–O entries of the currently active context. Correct responses produced a non-zero outcome (either $25 or $5, according to a pre-specified outcome structure), whereas incorrect responses produced outcome $0.

Sessions consisted of up to 320 trials and were organized into 10–16 blocks (randomly determined per session). Block lengths were randomly sampled between 15 and 32 trials. The two contexts alternated across blocks, inducing implicit context switches at block boundaries. The relationship between contexts was constructed as a global reversal: when the context switched, the correct response for every stimulus swapped between left and right (pairwise inversion).

## LLM evaluation

We evaluated a set of open-source large language models (LLMs): meta-llama/Llama-3.2-1B, meta-llama/Llama-3.2-3B, meta-llama/Llama-3.1-8B, meta-llama/Llama-3.1-70B, Qwen/Qwen2.5-0.5B, Qwen/Qwen2.5-1.5B, Qwen/Qwen2.5-3B, Qwen/Qwen2.5-7B, Qwen/Qwen2.5-14B, Qwen/Qwen2.5-32B, and Qwen/Qwen2.5-72B. All models were loaded using HuggingFace Transformers and evaluated in a pure inference setting with no parameter updates or fine-tuning.

For each model, we generated 30 independent sessions using the procedure above. We interfaced with LLMs using a sequence-completion paradigm in which the entire session history was represented as line-delimited text, one line per trial. Each line followed the format “stimulus response outcome”, where response and outcome for completed trials were the ground-truth environmental feedback included in the prompt.

To read out the model’s decision on trial $i$, we appended only the current trial’s stimulus token(s) to the prompt prefix and did not append that trial’s response or outcome. We then extracted the model’s next-token logits and restricted the choice set to {left, right}; the candidate with higher logit was taken as the model’s predicted response. After recording the prediction, we appended the “response outcome” for trial $i$ to the prompt as feedback, so that subsequent predictions were conditioned on the full history.

We computed an outcome from prediction correctness: if the predicted response matched the ground-truth response, the outcome was set to the trial's designated non-zero reward ($25 or $5); otherwise the outcome was set to $0. We logged trial-by-trial predicted responses for each session and recorded this derived outcome alongside the ground-truth response/outcome feedback used in the prompt. Finally, we saved complete per-trial session logs for all models for downstream Behavioural analyses.

**Human study**

Human Behavioural data were collected via a web-based experiment deployed on Amazon Mechanical Turk (MTurk). Each participant completed exactly one session (one session per subject). The human task used the same pseudo-word stimulus scheme and the same randomized session generation procedure as in the LLM evaluation, including the trial limit (up to 320), block counts (10–16), block length range (15–32), blockwise alternation of the two latent contexts, global left/right inversion across contexts, and the same outcome structure (correct: $25 or $5; incorrect: $0).

On each trial, participants viewed one of the four pseudo-word stimuli and selected a response (left or right) using either keyboard arrow keys or on-screen buttons. Immediately after each response, participants received outcome feedback: correct responses yielded $25 or $5 and incorrect responses yielded $0. The task proceeded trial-by-trial until the session ended.

To ensure data quality, a 4-digit completion code required for MTurk submission was displayed only if overall session accuracy was at least 80%. Sessions below this threshold did not produce a completion code and were excluded from the main analyses. In total, we recorded 58 sessions; 17 were excluded for failing the inclusion criterion, leaving 41 included sessions for downstream analyses.

**Behavioural analysis**

We performed quality control and grouping on the Behavioural sequence of each session, with two primary goals: (i) to exclude invalid sessions that failed to establish stable stimulus–response learning; and (ii) among valid sessions, to distinguish whether subjects/models exhibited inference Behaviour—namely, switching rules based on a single negative feedback signal. To do so, we first partitioned the trial sequence into blocks according to context switches, and within each block identified trials corresponding to the first occurrence of any stimulus within that block. Such trials capture the decision made when a subject/model encounters that stimulus for the first time under a new context, and therefore provide a direct test of whether a previously learned mapping structure is transferred to unseen stimuli.

Invalid sessions were excluded based on stable accuracy in non-inference conditions. Specifically, we selected control trials most relevant to the criterion "the rule has been learned," namely trials immediately preceding each context switch, where the subject/model remained in the old context and should have already mastered the mapping. We then tested whether the overall accuracy on these control trials was significantly above chance (0.5; one-sided binomial test, threshold $P <$ 0.05). If this test was not significant, we considered the session not to have reached a reliable level of stimulus–response learning; subsequent inference classification would therefore be uninterpretable, and the session was removed from Behaviour analysis and downstream geometric analyses.

For sessions passing the above screening, we further classified them as inference-present or inference-absent based on performance on inference trials. We defined inference trials as the first-occurrence trials of a given stimulus after a context switch, we excluded the earliest one, which typically provides the "switch cue" (and may yield the first negative feedback). Subsequent first-occurrence trials are better suited to assess whether the subject/model performs inference in the sense of "after observing a single piece of evidence, switching the entire stimulus–response mapping to the alternative context." We performed a one-sided binomial test (against 0.5) on the accuracy of this first inference trial aggregated across all blocks. If accuracy was significantly above chance ($P < 0.05$), we labeled the session as inference-present; otherwise, inference-absent. Thus, all included sessions exhibited reliably high accuracy in non-inference conditions, and were then grouped according to whether inference-trial performance was significantly above chance.

**Representational geometry analysis**

We performed forward passes of each LLM on the task sequences and extracted hidden representations at two time points per trial for geometric analyses: a baseline stage and a stimulus stage. The baseline-stage representation was defined as the hidden state of the last token in the context immediately before the trial's stimulus was presented. The stimulus-stage representation was defined as the hidden state at the position after the stimulus had been written into the context and immediately before the model produced its response. These two stages decompose "task-relevant representations" into an internal state without stimulus input and an immediate stimulus-driven representation. The baseline-stage state most directly carries trial-independent task variables, such as the current belief about context, updates induced by recent feedback, and a strategy state used to guide the next decision. The stimulus-stage state is the direct input to response generation; its geometry simultaneously reflects stimulus-identity drive and modulation by context/rule state.

In geometry analyses, we treated baseline- and stimulus-stage representations separately. To avoid introducing non-representational components into the geometric results, we excluded the embedding layer (layer 0) from all analyses, because it primarily reflects static token input encodings rather than task-relevant states formed under the current context. We also excluded the final layer, whose representations are typically closest to the output word-distribution (logit) readout space and thus strongly constrained by the output head and vocabulary geometry, which can obscure more general computational representational structure in intermediate layers. We therefore used only intermediate transformer-layer outputs for representational geometry analyses.

For each session, we used a unified preprocessing pipeline. Unless otherwise noted, we adopted a merged-layer representation that concatenates each trial's hidden representations across intermediate transformer layers into a single vector, followed by L2 normalization to remove overall magnitude differences; this vector was then used as the model representation at that time point for downstream geometric analyses. We then combined representations with the session's Behavioural records, retained correct trials only, and assigned each trial a condition label based on the task variables (context, response, outcome), enabling downstream analyses on the concatenated overall representation.

To reduce feature dimensionality and improve estimation stability, we optionally applied PCA separately to each layer (or to the merged representation produced by merge_layer). We first fitted PCA on the trial representations within the session, and automatically selected the minimum number of principal components needed to reach a cumulative explained-variance threshold (default 0.95). We then projected representations into the resulting low-dimensional subspace, which served as input for decoding and geometric-metric computation.

After preprocessing, we partitioned trials into 8 conditions (2×2×2) according to (context, response, outcome) and organized conditions in a fixed order. If a session did not cover all 8 conditions, it was excluded from geometry analyses. For included sessions, we sampled a fixed number K of trials within each condition (default K=15). When a condition contained fewer than K trials, we used sampling with replacement to balance sample sizes across conditions. This sampling procedure was repeated multiple times (default 10 repeats), and results were averaged across repeats to obtain stable estimates.

**Geometric metrics**

All geometric metrics were computed separately for each of the 35 balanced binary dichotomies (4 vs 4) obtained by enumerating all possible partitions of the 8 conditions. For each resampling instance, we first grouped the sampled trials into 8 condition-specific feature sets. Given a dichotomy, we concatenated features from the positive and negative condition sets to form a feature matrix $X$ and binary labels $y \in \{+1, -1\}$.

Decoding accuracy was estimated as the prediction accuracy of a linear SVM under 5-fold cross-validation. Before training, each feature dimension was z-scored (subtracting the mean and dividing by the standard deviation) using training data only; we then obtained out-of-fold predictions across the 5 folds and computed the cross-validated accuracy. The corresponding null decoding accuracy was obtained by randomly permuting the class labels $y$ and repeating the same cross-validation procedure, controlling for chance separability arising from finite samples and high-dimensional features.

CCGP (cross-condition generalization performance) quantifies the ability of a linear decision boundary to generalize across condition combinations. For each dichotomy, we performed a "3/1 holdout" cross-condition evaluation: among the 4 positive conditions, trials from 3 conditions were used for training and the remaining 1 condition for testing; the same was done on the negative side, resulting in a test set composed of one held-out positive condition and one held-out negative condition. We enumerated all train/test splits on both sides (4 choices per side; 16 combinations in total). For each split, we z-scored each feature dimension using the training set (subtracting the training mean and dividing by the training standard deviation), trained a linear SVM on the training set, and computed accuracy on the test set. The CCGP for a dichotomy was defined as the average accuracy across all 16 split combinations.

PS (parallelism score) measures whether encoding directions are consistent across different condition pairs within the same dichotomy. We first computed the mean feature vector $\mu_c$ for each of the 8 conditions. For a given dichotomy, we considered all perfect matchings between the 4 positive means and the 4 negative means ($4! = 24$ matchings). For each matching, we constructed four coding vectors (negative mean minus its matched positive mean), normalized each vector to unit length, and computed the average pairwise cosine similarity among the four vectors. PS was defined as the maximum of this average similarity over the 24 matchings.

All three metrics were computed independently on baseline and stimulus representations. After repeated resampling, we obtained a stable mean estimate for each session, which served as the unit of analysis for subsequent cross-session comparisons (e.g., inference-present vs. inference-absent) and statistical testing.

**Layer-wise geometric analysis**

To quantify how representational geometry evolves with depth, we repeated the geometric analyses described above separately for each intermediate transformer layer (the same layer-exclusion rules apply). Concretely, instead of forming the merged-layer vector, we used a per-layer representation in which each layer's hidden state was L2-normalized and then passed through the same downstream pipeline (trial filtering, (context,response,outcome) condition assignment, PCA, balanced within-condition resampling, and metric computation). This yielded, for each session and each time point (baseline or stimulus), layer-wise estimates of decoding accuracy, cross-condition generalization performance (CCGP), and parallelism score (PS).

Because models differ in depth, we mapped each layer index to a relative depth $d \in (0,1]$ by dividing by the maximum layer index of that model. For cross-model aggregation, each session's layer-wise metric curve was interpolated onto a shared relative-depth grid (50 equally spaced points by default). Within each session, metrics were first averaged across resampling repeats to obtain a stable estimate at each depth. When reporting task-variable–specific components (e.g., context, response, outcome, or selected composite dichotomies), we aggregated entries of the layer-wise dichotomy vector according to a pre-specified index set.

At the model level, we averaged interpolated session curves to obtain a model-specific depth trend. At the group level, we averaged model curves to obtain an overall depth profile and visualized uncertainty using standard error of the mean (s.e.m.) bands across models. Baseline- and stimulus-stage depth trends were computed and reported separately.

**Block-wise geometric analysis**

To obtain a coarse-grained view of how representational geometry differs across depth while avoiding noisy per-layer fluctuations, we performed a depth-block analysis that aggregates intermediate layers into a small number of contiguous blocks (default $n_{blocks} = 3$, corresponding to lower/middle/higher depth). Building on the preprocessing and trial/condition handling described in Representational geometry analysis, we first split the (included) intermediate transformer layers into $n_{blocks}$ near-equal contiguous segments. For each trial and each block, we concatenated the hidden representations of all layers within that block into a single vector and applied L2 normalization (block-wise "merge-block" representation). This produced, for every session, a set of block-specific representations $\{h^{(b)}\}_{b=1}^{n_{blocks}}$, which were then analyzed independently.

For each depth block $b$, we applied the same downstream geometry pipeline as in the main analysis—condition assignment by (context, response, outcome), balanced within-condition resampling, PCA, and computation of decoding accuracy, CCGP, and PS—yielding block-specific metric estimates per session. We then compared inference-present versus inference-absent sessions within each block, and summarized group effects using bootstrap aggregation over sessions (with matched null estimates computed using the same session-level procedure).

**Visualization**

To provide an intuitive view of representational geometry, we visualized sessions after grouping them by predefined criteria (e.g., inference-present vs. inference-absent; finetuned vs. non-finetuned). For each session, we first aggregated trial-level representations from the 8 conditions (defined by context × response × outcome), excluding incorrect trials, and computed a mean vector for each condition. We then averaged these condition means across sessions within each group, yielding 8 group-level condition centroids per group.

To enable direct comparison between two groups, we defined a shared orthogonal coordinate system using the union of the two groups' data, and projected both groups' condition centroids into this common space. The coordinate system consists of three interpretable geometric axes. The context axis is defined as the difference between the mean representations of the two contexts. Under the constraint of orthogonality to the context axis, we further defined a response axis (difference between left vs. right means) and an outcome axis (difference between the two outcome means), resulting in a consistent and interpretable 3D embedding across groups.

In the 3D plots, we used dashed lines to connect the positions of the same stimulus across the two contexts, thereby visualizing the cross-context displacement structure of identical stimuli. We also rendered semi-transparent surface patches for the four vertices within each context to illustrate the relative geometric relationship between the two contexts in this shared space. Both groups were plotted with identical axis ranges and camera/view parameters to ensure that visual comparisons of geometric shape were not confounded by differences in scaling or rotation. For interactive inspection, we displayed paired 3D views side-by-side during plotting.

**Sequence-prediction fine-tuning**

Improving a model's learning and inference performance on the implicit context-switch task via parameter updates may coincide with systematic improvements in representational geometry. If both strengthen synchronously after fine-tuning, this would provide more direct evidence for an association between representational geometry and LLM generalization performance. To test this, we performed supervised sequence-based fine-tuning on base LLMs using LoRA (parameter-efficient fine-tuning; PEFT), updating only a small set of low-rank adapter parameters while keeping the base weights unchanged.

Training data were synthesized using the same task-generation procedure as in evaluation. Each training example corresponds to a full session-level sequence of trials, written line-by-line as "stimulus response outcome", where both response and outcome are the trial-level ground truth. We split training and validation sets at the stimulus-string level: a subset of tokens in the pseudoword pool was held out exclusively for validation, ensuring that validation performance reflects generalization to unseen stimulus labels rather than memorization. We ensured that training, validation, and evaluation were strictly disjoint not only in stimulus labels but also in session realizations. We used non-overlapping pseudo-word pools for train/val/evaluation, so that the stimulus strings appearing in each split come from different token sets. In addition, even when using the same task-generation procedure, sessions were re-sampled independently for each split: the trial sequences within a session (including trial order and the stochastic realization of the task) were generated with fresh randomness, so that no session-level trial trajectory is reused across splits. This design prevents leakage at both the stimulus-identity level and the session-trajectory level, and ensures that validation/evaluation reflect generalization beyond memorization of specific pseudo-word labels or particular trial sequences.

Sequence-prediction fine-tuning used sparse token-level supervision: we computed the language-modeling loss only at positions directly relevant to Behavioural decisions, while masking all other token positions (excluded from the loss). Specifically, supervision covered the key tokens corresponding to the trial's response and outcome, but imposed no prediction constraint on inter-trial newline tokens or the stimulus token itself. To avoid training on the "evidence trial" immediately following a context switch (which typically contains negative feedback and triggers subsequent inference updates), we also masked the response/outcome labels for that trial, and applied supervision only on segments where context is stable. Fine-tuning used the standard causal language modeling objective (maximum likelihood; next-token prediction), minimizing cross-entropy over the unmasked label positions. Optimization used AdamW; for reproducibility and downstream inference we saved and loaded only the LoRA adapter parameters. During training, we periodically evaluated inference-trial accuracy on the validation set, using this metric to track progress and select the best-performing checkpoint.

**Geometry-constrained fine-tuning**

Our preliminary results indicate that a model's generalization ability is closely related to the geometric structure of its internal representations. Motivated by this, we further asked whether directly optimizing representational geometry can improve inference-like Behaviour on the implicit context-switch task, providing evidence for a more practically meaningful training paradigm: enhancing generalization by shaping representational structure, without relying on additional annotations or task-specific rule engineering. To test this, we performed geometry-constrained fine-tuning. Training examples remained full session sequences (trials concatenated as “stimulus response outcome”), but in this setting we did not optimize the standard language-modeling loss. Instead, we applied only a geometry-constraint loss to enforce structure on the representations; observable Behavioural performance (e.g., validation inference-trial accuracy) was computed separately during evaluation.

The geometry-constraint loss was constructed within each training batch from hidden representations at the stimulus time point. We first aggregated layer-wise hidden states from the forward pass, and for each trial selected the token position aligned to the trial’s stimulus as that trial’s stimulus representation. We then assigned stimulus representations to the 8 conditions in the 2×2×2 factorial design according to the ground-truth triplet (context, response, outcome). Within each batch, we averaged representations that share the same triplet to obtain a representative vector for each condition.

We then shaped abstract geometry by enforcing consistency of three classes of difference-coding directions: (i) context differences, computed between contexts while holding response and outcome fixed; (ii) response differences, computed between responses while holding context and outcome fixed; and (iii) outcome differences, computed between outcomes while holding context and response fixed. For each class, we required that difference vectors derived under different “background combinations” (i.e., different assignments of the other two variables) be as parallel as possible in representation space. Concretely, the loss penalized deviations from parallelism using pairwise cosine similarity among difference vectors, and summed the penalties across the three classes to form the final geometry-constraint loss. This objective directly encourages context, response, and outcome to form reusable, directionally consistent abstract coding axes across conditions, thereby increasing compositionality and cross-condition generalization structure.

Training used LoRA adapters to update only low-rank parameters while freezing the base model. When geometry constraints were enabled, we disabled the default cross-entropy language-modeling objective and optimized exclusively the geometry-constraint loss. Importantly, we implemented this loss in its most minimal form, without additional engineering (e.g., complex sampling schemes, hard-negative mining, dynamic weighting, or multi-stage training). The goal was to test as directly as possible whether imposing simple, interpretable structural constraints on representational geometry alone is sufficient to alter abstract representations and inference-like Behavioural performance.

**Null distributions for representational-geometry metrics**

We used different null constructions depending on the metric, matching the computation pipeline while selectively destroying the geometry that supports abstraction.

(1) Label-shuffle null for SD (chance decoding under identical data).

For SD, after constructing the decoding dataset for a given dichotomy (concatenating the sampled trials from the positive vs. negative condition sets), we generated a null score by shuffling the class labels $y$ and recomputing cross-validated linear-SVM accuracy on the same features. This preserves the full representational distribution and any within-condition covariance, while enforcing the null hypothesis of no relationship between labels and representations.

(2) "Geometric null" for CCGP and PS (dimension permutation within condition).

For each of the 8 condition-specific matrices $X_c \in \mathbb{R}^{K \times D}$, we independently applied a random permutation of the feature dimensions, yielding

$$X_c^{(0)} = X_c[:, \pi_c],$$

where $\pi_c$is a random permutation of $\{1, \dots, D\}$ drawn per condition. This operation preserves the per-dimension marginal statistics and trial structure within each condition, but breaks consistent coding directions across conditions that would support cross-condition generalization or parallelism. We then recomputed CCGP and PS on these permuted representations using the same train/test splits, standardization, and classifier settings as for the observed data, producing a null value per dichotomy.

**Session-level aggregation, uncertainty estimation, null distributions and permutation testing**

All geometry metrics were treated as session-level quantities to avoid pseudo-replication across trials. For each session $s$, we obtained a vector of metric values $m_s \in \mathbb{R}^D$(with $D$ dichotomies). Because metric estimation involves stochastic trial subsampling (and, for some metrics, stochastic resampling within the pipeline), we repeated the session-level computation $R$ times and averaged across repeats to reduce Monte-Carlo noise:

$$m_s = \frac{1}{R} \sum_{r=1}^{R} m_{s,r}.$$

Unless stated otherwise, we used $R = 50$ repeats per session.

For each group $g$ (e.g., inference-present or inference-absent), we aggregated sessions by a simple unweighted mean:

$$\bar{m}_g = \frac{1}{N_g} \sum_{s \in g} m_s,$$

where $N_g$is the number of valid sessions in the group. All reported group means (and downstream contrasts) therefore reflect equal weighting of sessions.

To quantify uncertainty of group-level estimates, we used a nonparametric bootstrap over sessions. For each bootstrap draw $b = 1, \dots, B$, we sampled $N_g$sessions from group $g$ with replacement and recomputed the group mean:

$$\bar{m}_g^{(b)} = \frac{1}{N_g} \sum_{i=1}^{N_g} m_{s_i^{(b)}} \, .$$

The resulting bootstrap distribution $\left\{\bar{m}_g^{(b)}\right\}_{b=1}^{B}$ was used to form 95% percentile confidence intervals (2.5th–97.5th percentiles). We used $B = 1000$ bootstrap draws for group means.

For between-group contrasts, we reported the difference in group means, $\Delta = \bar{m}_{\text{present}} - \bar{m}_{\text{absent}}$, and estimated its uncertainty using a two-sample bootstrap: resampling sessions independently within each group, recomputing $\bar{m}_{\text{present}}^{(b)}$, $\bar{m}_{\text{absent}}^{(b)}$ and $\Delta^{(b)}$, then taking the 2.5th–97.5th percentiles of $\left\{\Delta^{(b)}\right\}$ as the 95% CI.

To compare observed group-level geometry against chance, we constructed a group-level null distribution by propagating session-level null repeats through the same session-aggregation step. For each session $s$, the session pipeline produced $R$ null repeats $m_{s,r}^{(0)}$. For each group-level null draw $b$, we independently sampled a repeat index $r_s^{(b)} \sim \text{Unif}\{1, \ldots, R\}$ for every session, and computed the mean across sessions:

$$\bar{m}_{g,0}^{(b)} = \frac{1}{N_g} \sum_{s \in g} m_{s, r_s^{(b)}}^{(b)}.$$

This yields a null distribution of group-mean metric values that matches the observed pipeline's aggregation and repeat structure. We used the same number of draws ($B = 1000$) to summarize null percentiles (e.g., 5th–95th bands) and to compute exceedance probabilities where needed (e.g., the fraction of null draws exceeding the observed group mean, with a +1 correction for finite-sample stability).

To test between-group differences, we additionally performed a session-level permutation test. For each metric, analysis stage and dichotomy, the observed test statistic was the difference in group means, $\Delta = \bar{m}_{\text{present}} - \bar{m}_{\text{absent}}$. We randomly permuted inference present and inference absent labels across sessions while preserving the original group sizes, recomputed $\Delta$ for each permutation, and estimated a two-sided empirical P value as the fraction of permuted $|\Delta|$ values greater than or equal to the observed $|\Delta|$, with a +1 correction for finite-sample stability. Unless otherwise stated, 10,000 permutations were used.

## Extended Data Figures

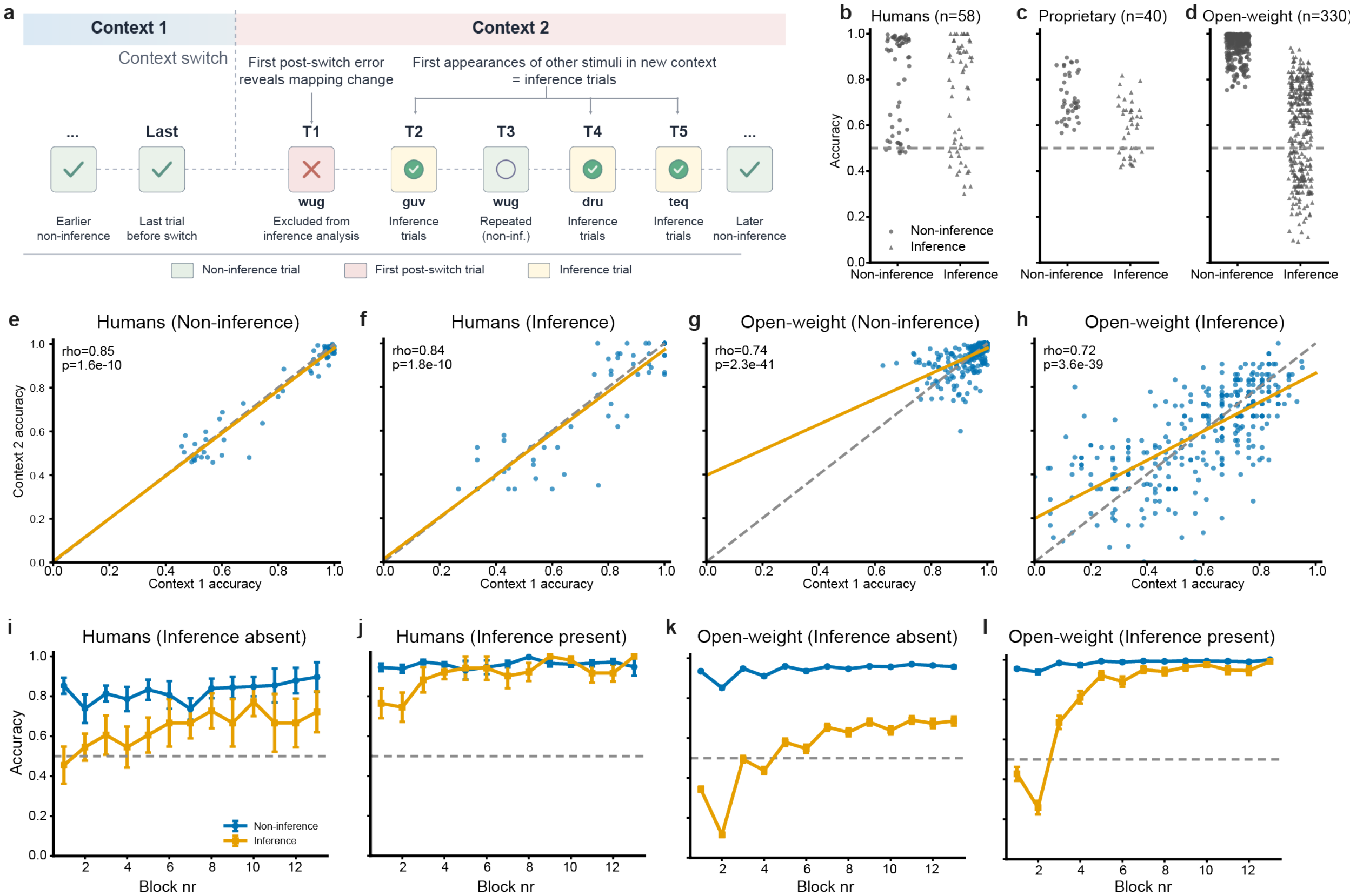

**Extended Data Fig. 1 | Behavioural definitions and statistics.** (a) Definition of inference and non-inference trials. Following an implicit context switch, the first post-switch error trial reveals that the previously valid mapping no longer holds and is excluded from the inference analysis. In this schematic, repeated presentations of the same stimulus are also excluded. Inference trials are defined as the first appearances of the other stimuli in the new context, whereas subsequent trials are classified as non-inference trials. (b-d) Session-level separation of non-inference and inference performance across datasets. Behavioural sessions from humans (n=58) (d), proprietary models (n=40) (e), and open-weight models (n=330) (f) were analyzed in a two-context reversal-learning task. Trials were partitioned into non-inference trials and inference trials (the second to fourth first-occurrence trials after each context switch). For each session, mean accuracy was computed separately for the two trial classes and plotted as individual points, with a dashed reference at chance level (0.5). (e-h) Cross-context consistency of session accuracy for non-inference and inference trials. For each session, accuracy was computed independently in context 1 and context 2 for non-inference and inference trials. Context 1 accuracy was plotted against context 2 accuracy in separate panels, with the identity line indicating perfect cross-context agreement. Spearman rank correlation (rho, p) quantified cross-context correspondence, and a least-squares fit line summarized the linear trend. (i-l) Blockwise accuracy dynamics stratified by inference expression. Sessions were classified as inference-present or inference-absent using right-tailed binomial tests against chance (p=0.5): baseline (last-trial) accuracy had to exceed chance (p<0.05), and inference-trial accuracy was required to be above chance for the present group (p<0.05) or not above chance for the absent group (p>0.05). Within each group and dataset, block-level accuracy was computed separately for non-inference and inference trials,

then aggregated across sessions as mean ± s.e.m. Curves are shown over the first 13 blocks with a horizontal chance reference line.

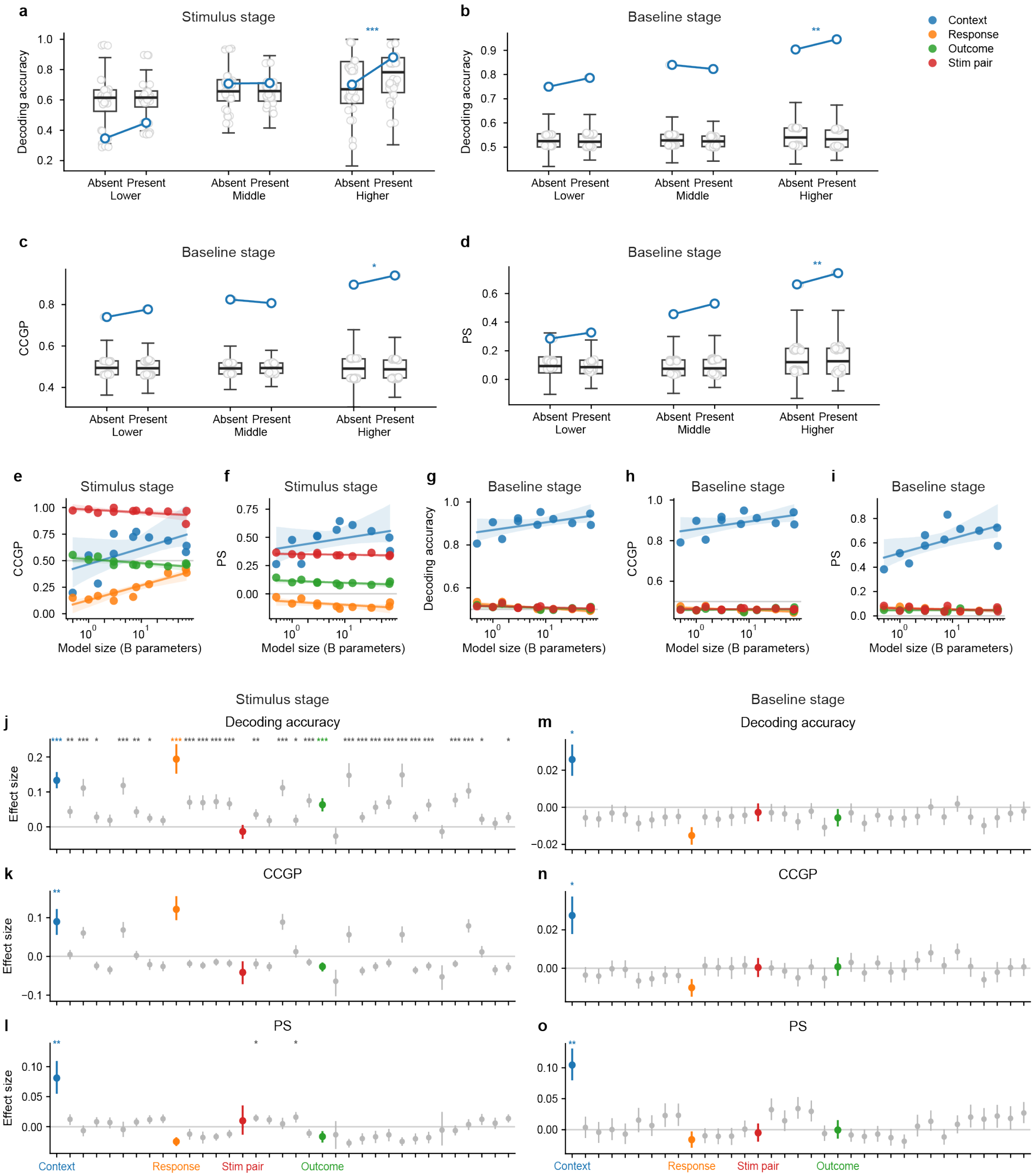

**Extended Data Fig. 2 | Layer-block, model-scaling and complete dichotomy-level statistics for inference-related representational geometry.** a, Stimulus-stage layer-block analysis of decoding accuracy. Plotting conventions and significance criteria follow Fig. 2. Context decoding increased most strongly in higher layers (mean present/absent = 0.881/0.701; Δ = +0.180, CI95 [+0.146, +0.215], q = $1.75 \times 10^{-4}$), with a smaller effect in lower layers (0.449/0.346;

$\Delta = +0.103$, CI95 [+0.053, +0.152], $q = 1.08 \times 10^{-3}$) and no significant effect in middle layers (0.711/0.708; $\Delta = +0.004$, CI95 [−0.016, +0.024], $q = 0.871$). b–d, Baseline-stage layer-block analysis for decoding accuracy (b), cross-condition generalization performance (CCGP; c) and parallelism score (PS; d). In higher layers, context was already organized before stimulus insertion and was further strengthened in inference-present sessions for decoding accuracy (0.946/0.904; $\Delta = +0.042$, CI95 [+0.031, +0.053], $q = 6.99 \times 10^{-3}$), CCGP (0.940/0.896; $\Delta = +0.044$, CI95 [+0.032, +0.056], $q = 1.05 \times 10^{-2}$) and PS (0.742/0.663; $\Delta = +0.079$, CI95 [+0.051, +0.106], $q = 6.99 \times 10^{-3}$). Lower- and middle-layer baseline effects were weaker and did not survive correction across all three metrics. e–i, Cross-model scaling of representational geometry across open-weight models. Stimulus-stage CCGP (e) and PS (f) are shown together with baseline-stage decoding accuracy (g), CCGP (h) and PS (i). At the stimulus stage, context CCGP and PS showed positive but non-significant model-size trends (CCGP: Spearman $\rho = +0.492$, $P = 0.124$; PS: $\rho = +0.396$, $P = 0.228$; $n = 11$ models). At the baseline stage, context PS increased significantly with model size ($\rho = +0.656$, $P = 0.028$), whereas baseline context decoding accuracy and CCGP showed weaker positive trends (both $\rho = +0.346$, $P = 0.297$). j–o, Complete dichotomy-level effect statistics for all 35 balanced dichotomies. Panels show the inference-present minus inference-absent difference ($\Delta$) with bootstrap 95% confidence intervals for stimulus-stage decoding accuracy (j), CCGP (k) and PS (l), and for baseline-stage decoding accuracy (m), CCGP (n) and PS (o). In the stimulus stage, context increased across all three metrics: decoding accuracy (0.861/0.728; $\Delta = +0.133$, CI95 [+0.104, +0.162], $q = 3.50 \times 10^{-4}$), CCGP (0.666/0.576; $\Delta = +0.090$, CI95 [+0.049, +0.130], $q = 5.25 \times 10^{-3}$) and PS (0.549/0.468; $\Delta = +0.081$, CI95 [+0.048, +0.114], $q = 8.75 \times 10^{-3}$). Other named variables showed dissociable effects: response became more readable but less parallel, outcome CCGP and PS decreased, and stimulus pair remained near ceiling in decoding without a significant decoding change. In the baseline stage, context again showed the most consistent positive effect, with increased decoding accuracy (0.921/0.895; $\Delta = +0.026$, CI95 [+0.015, +0.036], $q = 1.93 \times 10^{-2}$), CCGP (0.911/0.884; $\Delta = +0.027$, CI95 [+0.016, +0.039], $q = 4.20 \times 10^{-2}$) and PS (0.702/0.598; $\Delta = +0.104$, CI95 [+0.075, +0.135], $q = 3.50 \times 10^{-3}$). By contrast, response showed small decreases in decoding accuracy and CCGP, whereas outcome and stimulus pair showed no significant positive baseline-stage geometric enhancement. Together, these complete forest plots distinguish selective context-related abstract-geometry enhancement from broader changes in linear readability. All $\Delta$ values denote inference-present minus inference-absent differences. CI95 indicates bootstrap 95% confidence intervals. q values are Benjamini–Hochberg-corrected permutation-test P values across the 35 dichotomies. Null construction, resampling and statistical tests are described in Methods.

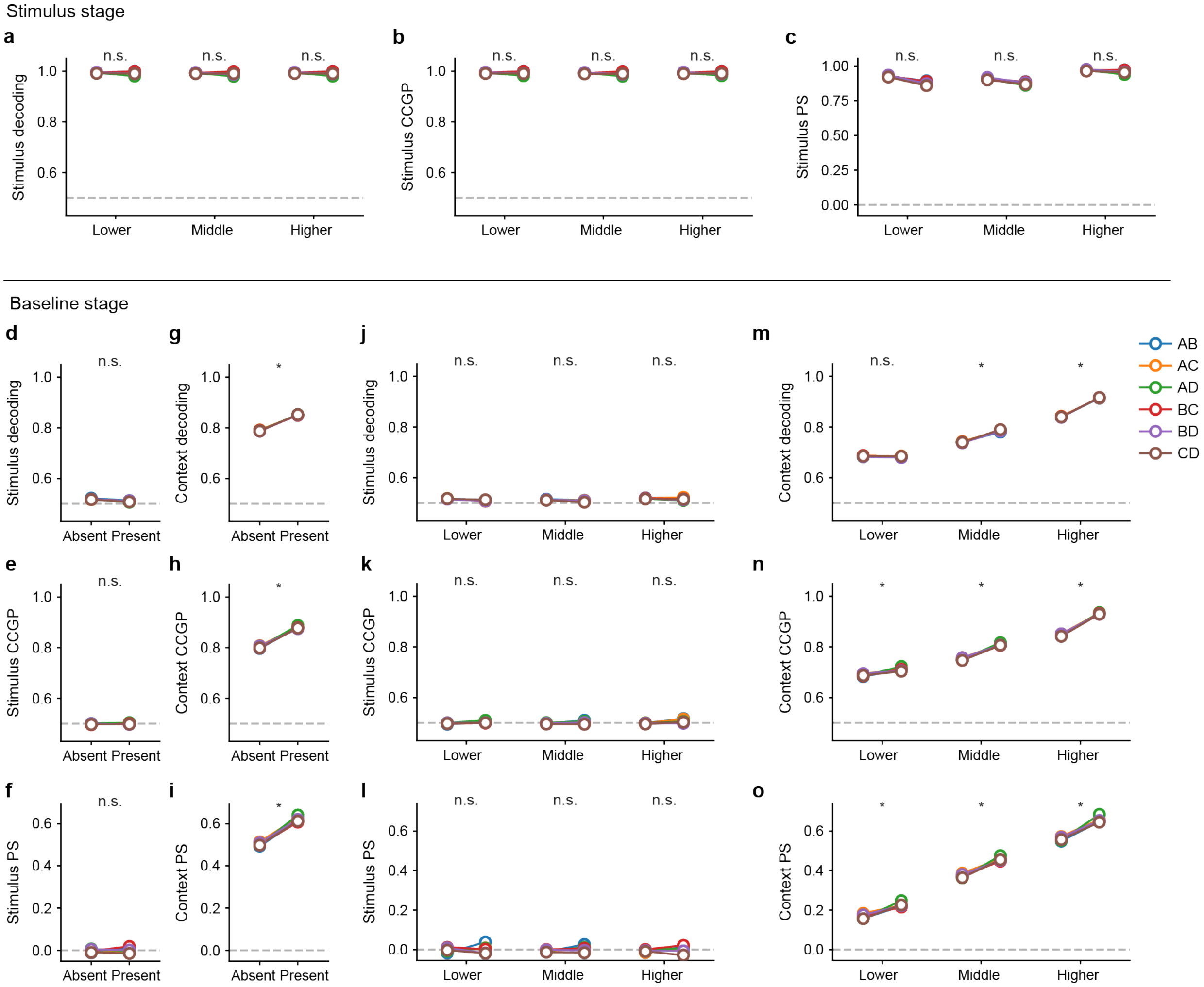

**Extended Data Fig. 3 | Stimulus identity remains stable, whereas baseline context geometry is strengthened across depth during inference.** a–c, Stimulus-stage stimulus-pair geometry across lower, middle and higher transformer blocks. For each stimulus pair, stimulus decoding (a), stimulus cross-condition generalization performance (stimulus CCGP; b) and stimulus parallelism score (stimulus PS; c) were computed after stimulus insertion and immediately before response generation. Open circles denote the six pairwise stimulus contrasts (AB, AC, AD, BC, BD and CD), with lines connecting inference-absent and inference-present sessions for the same contrast within each block. Pair-averaged stimulus decoding and CCGP were near ceiling in all three blocks and showed no positive inference-related effect (decoding, Present/Absent = 0.992/0.994, 0.991/0.993 and 0.991/0.994; CCGP = 0.992/0.994, 0.991/0.992 and 0.992/0.994 for lower, middle and higher blocks, respectively; all n.s.). Stimulus PS also showed no positive inference-related effect (Present/Absent = 0.876/0.929, 0.878/0.909 and 0.959/0.971; all n.s.). d–i, All layers baseline-stage stimulus and context geometry. Stimulus identity remained close to chance before stimulus arrival and showed no positive inference-related effect across stimulus decoding (d), stimulus CCGP (e) or stimulus PS (f). By contrast, latent context was already represented before stimulus arrival and was stronger in inference-present sessions across context decoding (g), context CCGP (h) and context PS (i; Present/Absent: context decoding = 0.794/0.755, context CCGP = 0.818/0.762 and context PS = 0.446/0.367). j–l, Depth-resolved baseline-

stage stimulus geometry. Stimulus decoding, stimulus CCGP and stimulus PS remained near chance or zero across lower, middle and higher blocks and showed no positive inference-related effect in any block (all n.s.). m–o, Depth-resolved baseline-stage context geometry. Context decoding increased across depth and was higher in inference-present sessions in the middle and higher blocks (lower: 0.683/0.685, Δ = −0.003, n.s.; middle: 0.785/0.740, Δ = +0.045, P = 0.0027; higher: 0.915/0.841, Δ = +0.075, P = 0.0028). Context CCGP was higher in inference-present sessions in all blocks (lower: 0.712/0.688, Δ = +0.023, P = 0.0026; middle: 0.810/0.751, Δ = +0.059, P = 0.0028; higher: 0.932/0.847, Δ = +0.085, P = 0.0022), as was context PS (lower: 0.226/0.167, Δ = +0.059, P = 0.0018; middle: 0.457/0.373, Δ = +0.084, P = 0.0028; higher: 0.656/0.560, Δ = +0.096, P = 0.0027). The grey dashed lines indicate chance level for decoding accuracy and CCGP, and zero for PS. Asterisks indicate permutation-test significance for pair-averaged inference-present versus inference-absent differences (*$P < 0.05$); n.s., not significant. Comparisons marked with asterisks additionally required the inference-present mean to exceed the inference-absent mean and the 95th percentile of the corresponding null distribution. Full resampling procedures, null construction and statistical tests are described in Methods.

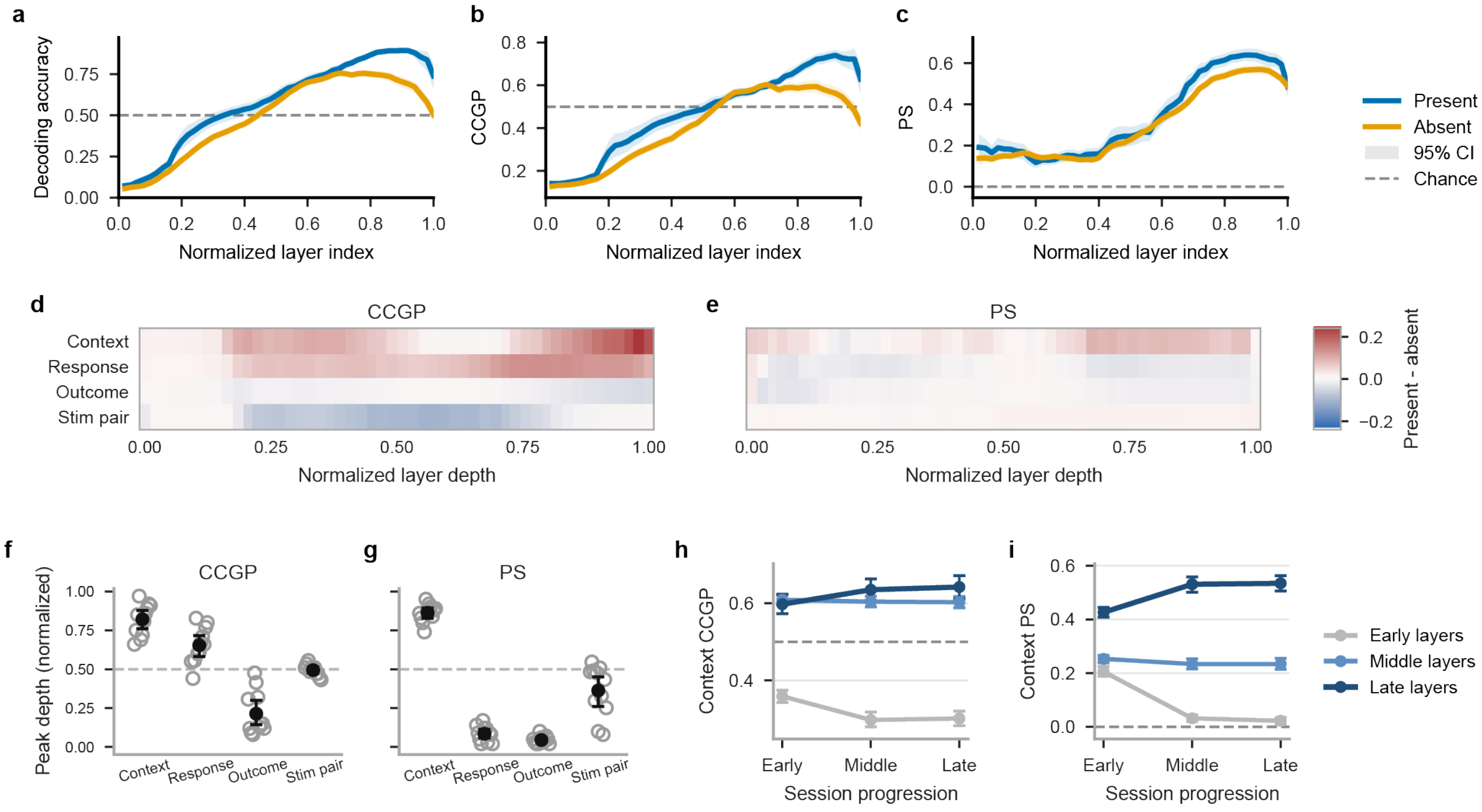

**Extended Data Fig. 4 | Layerwise emergence and blockwise evolution of geometry.** (a-c) Continuous layerwise profiles of context decoding accuracy (a), CCGP (b), and PS (c) across normalized layer index. Blue and orange curves denote inference-present and inference-absent sessions, respectively. Shaded bands indicate 95% bootstrap confidence intervals across sessions. Dashed lines mark the corresponding chance or baseline levels. (d-e) Layer-by-dichotomies heatmaps showing the difference between inference-present and inference-absent sessions across normalized layer depth for CCGP (d) and PS (e). Rows denote named dichotomies (context, response, outcome, and stim_pair), and columns denote normalized layer index. Color indicates the difference between the session-averaged values in inference-present and inference-absent groups. Context exhibits the most consistent enhancement across middle and higher layers, whereas other variables show weaker and more localized effects. (f-g) Peak-depth summary of layerwise

abstract geometry across open-weight models using CCGP (f) and PS (g). For each model, session-pooled layerwise curves were computed for context, response, outcome, and stim pair, and the peak layer position was identified after light smoothing. Peak positions are shown on a normalized depth scale from 0 to 1. Open circles denote individual models; black markers and error bars indicate the mean ± 95% bootstrap confidence interval across models. (h-i) Within-session progression of context geometry across progression bins defined by context switches. Each session was partitioned into relative progression bins (Early, Middle, Late) according to context switches, and context geometry was quantified separately for low-, middle-, and high-depth groups of layers. Context CCGP increases across progression bins primarily at high depth (h), with comparatively modest change at middle depth and decrease at low depth. Context PS shows a similar depth-dependent pattern (i), with the strongest rise again at high depth. Points denote cross-session means and error bars indicate 95% bootstrap confidence intervals. These panels indicate that experience-dependent reorganization of context geometry unfolds over session progression mainly in higher layers, where representations become both more generalizable across conditions and more geometrically parallel.

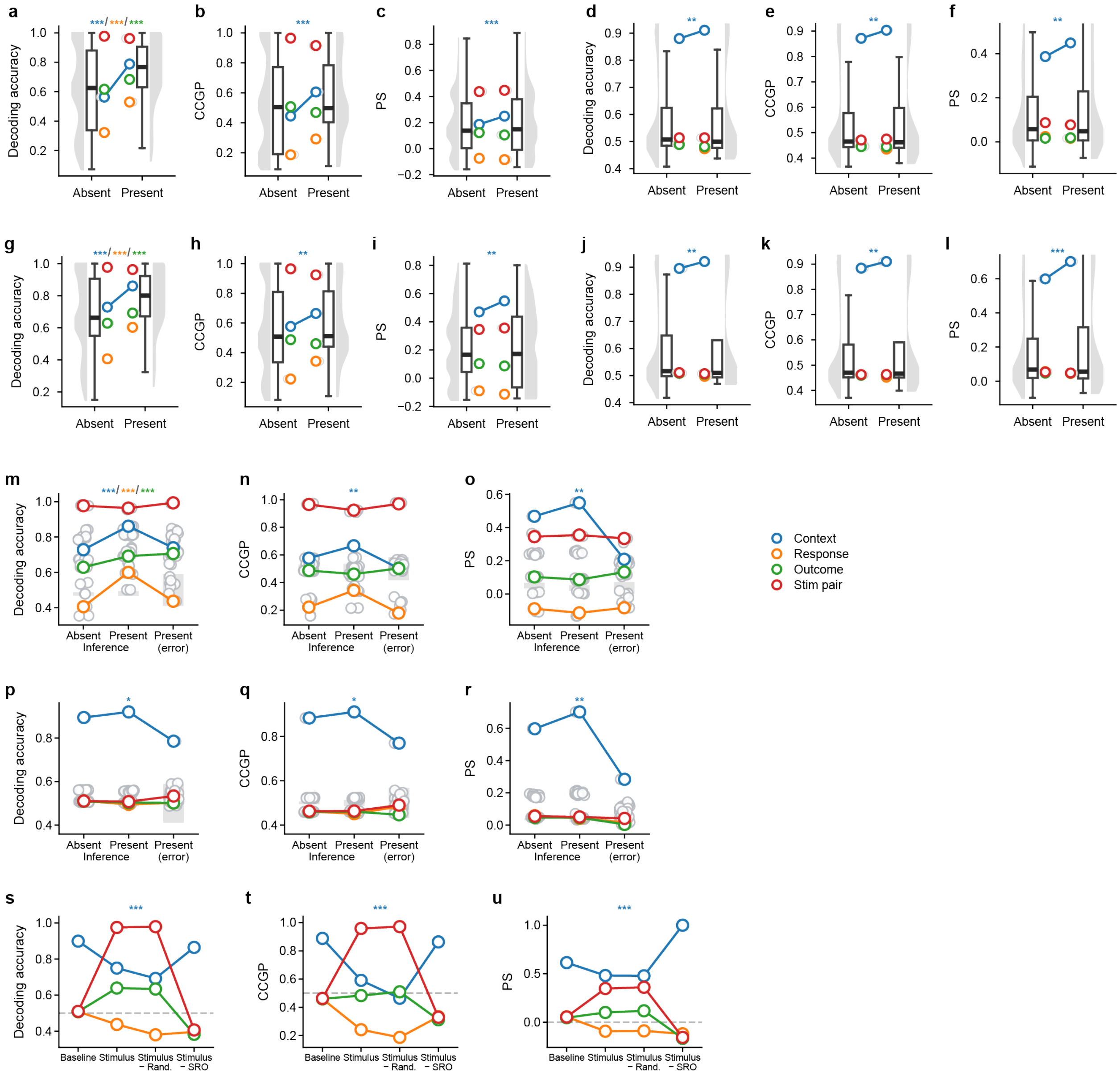

**Extended Data Fig. 5 | Robustness and trial-level specificity of inference-related abstract geometry.** (a–f) Geometry after joint removal of features with strong context, response or outcome main effects, with decoding accuracy, CCGP and PS shown for stimulus stage (a–c) and baseline stage (d–f) representations. (g–l) Geometry after representation-norm matching, with the same metrics shown for stimulus stage (g–i) and baseline stage (j–l) representations. (m–r) Error-trial analysis comparing correct trials from inference absent sessions, correct trials from inference present sessions and error trials from inference present sessions, shown for stimulus stage (m–o) and baseline stage (p–r) representations. (s–u) Stimulus stage subspace-control analysis after removing stimulus-, response- and outcome-related coding subspaces; raw stimulus stage, baseline stage and rank-matched random-removal values are shown for reference. Grey distributions or grey points summarize balanced dichotomies, and colored open circles mark the named dichotomies: context, response, outcome and stimulus pair. Stars are shown for named dichotomies only. In a–r, stars follow the same annotation criteria as in the main geometry analysis. In s–u, stars mark Holm-

adjusted one-sided paired permutation tests comparing stimulus stage values after SRO-subspace removal with raw stimulus stage values, shown only when the SRO-subspace-removed value also exceeded the corresponding metric baseline. Across the main-effect-removal and representation-norm-matched controls, inference present sessions retained elevated context CCGP and PS, indicating that the abstract context geometry was not driven by high main-effect features or global representation magnitude. The SRO-subspace analysis further showed that removing trial-specific stimulus, response and outcome components partially restored stimulus stage context CCGP and PS relative to raw stimulus stage values, whereas rank-matched random removal did not. This supports the interpretation that the stimulus stage preserves latent context but embeds it into a mixed decision-related representation. Conversely, context geometry was reduced on error trials, especially for PS, indicating that the abstract context axis is coupled to trial-level inference success.

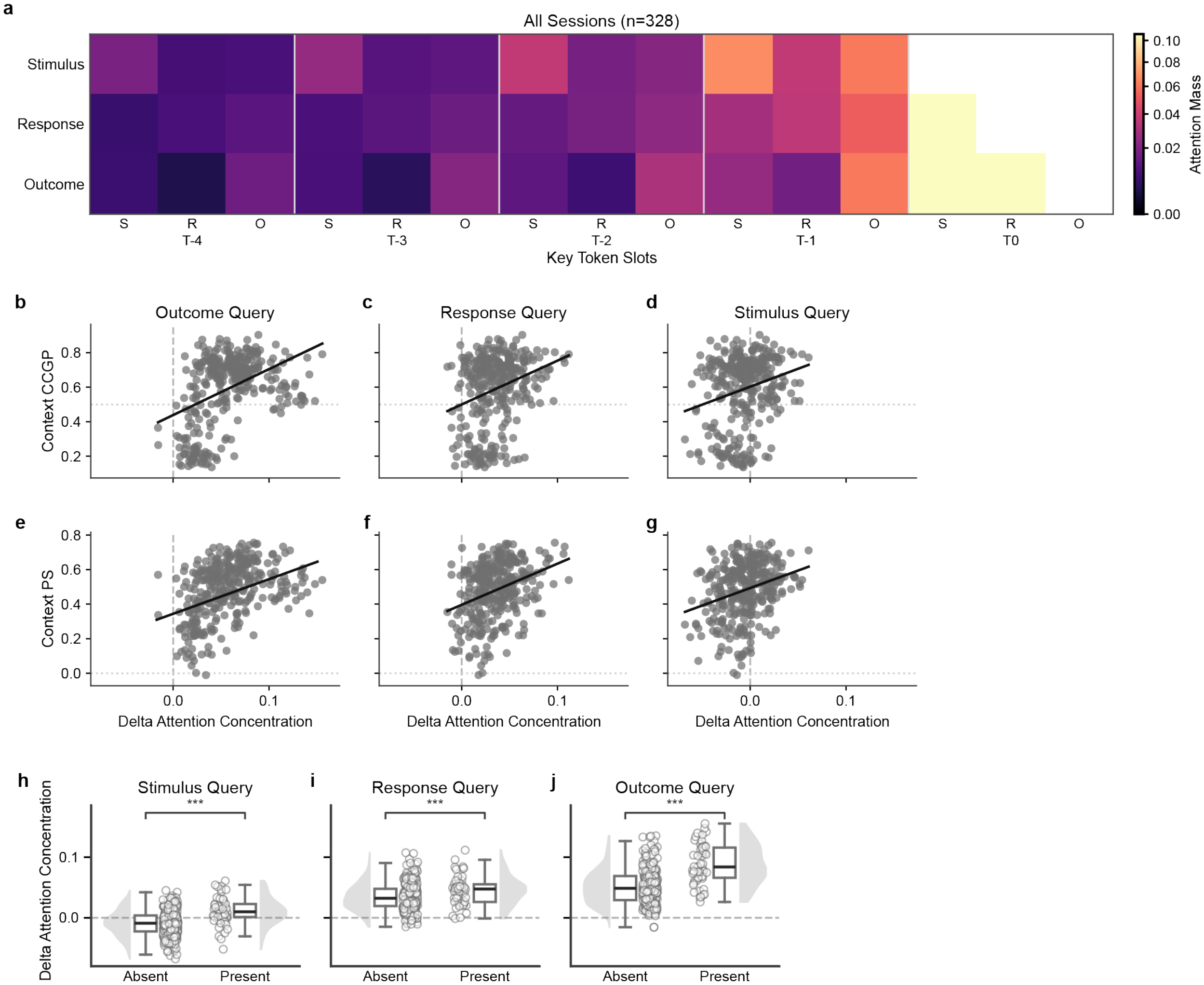

**Extended Data Fig. 6 | Attention reallocation is associated with abstract context geometry and inference success.** a, Mean attention distribution across all analyzed sessions (n = 328), resolved by queried S-R-O field and attended S-R-O field across relative trial positions T−4 to T0. Attention was concentrated on recent history and was query-specific, with distinct weighting of stimulus, response and

outcome fields across the three queried fields. Blank cells indicate unavailable current-trial fields. b–g, Session-level associations between attention reallocation around context switch and abstract context geometry. The x axis shows one of three predefined task-relevant changes in attention concentration: outcome-query concentration, post-switch minus pre-switch (b,e); response-query concentration, post-switch minus pre-switch (c,f); and stimulus-query concentration, first inference minus post-switch (d,g). The y axis shows context CCGP in b–d and context PS in e–g. Each dot denotes one session, black lines show linear fits, the vertical dashed line marks zero change in attention concentration, and the horizontal dashed line marks the geometry reference value, defined as chance level (0.5) for CCGP and the null reference (0) for PS. Attention reallocation was positively associated with both geometry metrics, with the strongest associations observed for outcome queries (b: Spearman $\rho = 0.395$, $q = 6.0 \times 10^{-5}$; c: $\rho = 0.252$, $q = 6.0 \times 10^{-5}$; d: $\rho = 0.195$, $q = 3.9 \times 10^{-4}$; e: $\rho = 0.440$, $q = 6.0 \times 10^{-5}$; f: $\rho = 0.320$, $q = 6.0 \times 10^{-5}$; g: $\rho = 0.273$, $q = 6.0 \times 10^{-5}$). q values in b–g were Holm–Bonferroni adjusted across the six Spearman correlation tests, corresponding to three query-specific attention deltas × two geometry metrics. h–j, The same three predefined attention-reallocation measures in inference-absent and inference-present sessions. Grey half-violins show the session distributions, box plots show the median and interquartile range, whiskers extend to 1.5× IQR, and open circles denote individual sessions. Dashed horizontal lines mark zero change. Inference-present sessions showed larger task-relevant attention reallocation than inference-absent sessions (h: $\Delta = 0.01987$, $q = 3.0 \times 10^{-4}$; i: $\Delta = 0.01142$, $q = 4.0 \times 10^{-4}$; j: $\Delta = 0.03692$, $q = 3.0 \times 10^{-4}$; two-sided permutation mean-difference test). q values in h–j were Holm–Bonferroni adjusted across the three predefined reallocation-delta comparisons. Together, these analyses show that structured attention reallocation is coupled both to stronger abstract context geometry and to behavioral inference success. All tests were two-sided. S, stimulus; R, response; O, outcome; T−4 to T0, relative trial positions from older to current trial. *$q < 0.05$, **$q < 0.01$, ***$q < 0.001$.